\documentclass[10pt]{ysupreprint}

\usepackage{booktabs}
\usepackage{multirow}
\usepackage{tabularx}
\usepackage{xspace}
\usepackage{enumitem}
\usepackage{amsmath}

\usepackage{amssymb}
\usepackage{mathtools}
\usepackage{tikz}
\usetikzlibrary{arrows.meta,positioning,fit,calc,shapes.geometric}

\newcommand{\systemname}{IntAct}
\newcommand{\system}{\textsc{\systemname}\xspace}

\usepackage{relsize}
\definecolor{codeground}{gray}{0.93}
\makeatletter
\newcommand{\code@box}[1]{{\setlength{\fboxsep}{1pt}\colorbox{codeground}{\vphantom{(gy}#1}}}
\newcommand{\code@space}{{\setbox\z@\hbox{\vphantom{(gy}\ }\color{codeground}%
  \leaders\hrule height\dimexpr\ht\z@+1pt\relax depth\dimexpr\dp\z@+1pt\relax
  \hskip\dimexpr\wd\z@-2pt\relax}}
\def\code@split#1 #2\@nil{%
  \def\code@rest{#2}%
  \ifx\code@rest\@empty
    \code@box{#1}%
  \else
    \code@box{#1}\code@space\code@split#2\@nil
  \fi}
\DeclareRobustCommand{\code}[1]{{\relscale{0.9}\ttfamily\code@split#1 \@nil}}
\makeatother

\newcolumntype{Y}{>{\raggedright\arraybackslash}X}

\tikzset{
  stage/.style={draw, rounded corners, align=center, minimum height=8mm, inner sep=4pt, fill=blue!5},
  action/.style={draw, rounded corners, align=center, minimum height=8mm, inner sep=4pt, fill=green!8},
  check/.style={draw, diamond, aspect=2.2, align=center, inner sep=2pt, fill=orange!10},
  flow/.style={-{Latex[length=2mm]}, thick},
  note/.style={align=left, font=\footnotesize}
}

\usepackage{pifont}
\usepackage{fontawesome5}   %
\usetikzlibrary{decorations.pathreplacing,shapes.misc,backgrounds}
\usepackage{subcaption}
\usepackage{float}
\usepackage{algorithm}
\usepackage{algpseudocode}

\newcommand{\iecbench}{\textsc{IEC-Bench}\xspace}
\newcolumntype{L}[1]{>{\raggedright\let\newline\\\arraybackslash\hspace{0pt}}m{#1}}
\newcolumntype{C}[1]{>{\centering\let\newline\\\arraybackslash\hspace{0pt}}m{#1}}
\newcolumntype{R}[1]{>{\raggedleft\let\newline\\\arraybackslash\hspace{0pt}}m{#1}}
\newcommand{\find}[1]{
\begin{tcolorbox}[leftrule=1mm,toprule=0mm,bottomrule=0mm,left=1pt,right=2pt,top=2pt,bottom=2pt]
\em #1
\end{tcolorbox}
}

\title{Do Tool Calls Execute as Intended? Measuring and Repairing Intent-Execution Correspondence in LLM Agents}

\authors{Boyang Yang\aff{1}, Zhenhao Li\aff{1}, Ziyao Yang\aff{1}, Kanghui Jia\aff{1}, Xin Yin\aff{2}, Mingmou Liu\aff{3}, Haoye Tian\aff{4}}
\affiliations{\aff{1}Yanshan University, \aff{2}Zhejiang University, \aff{3}Nanjing University, \aff{4}Aalto University}

\metadata{Date}{October 3, 2026}
\metadata{Correspondence}{Haoye Tian}
\metadata{Claude Code Mod}{\url{https://github.com/GLEAM-Lab/intact-bash-mod}}
\logo{\includegraphics[height=1.8cm]{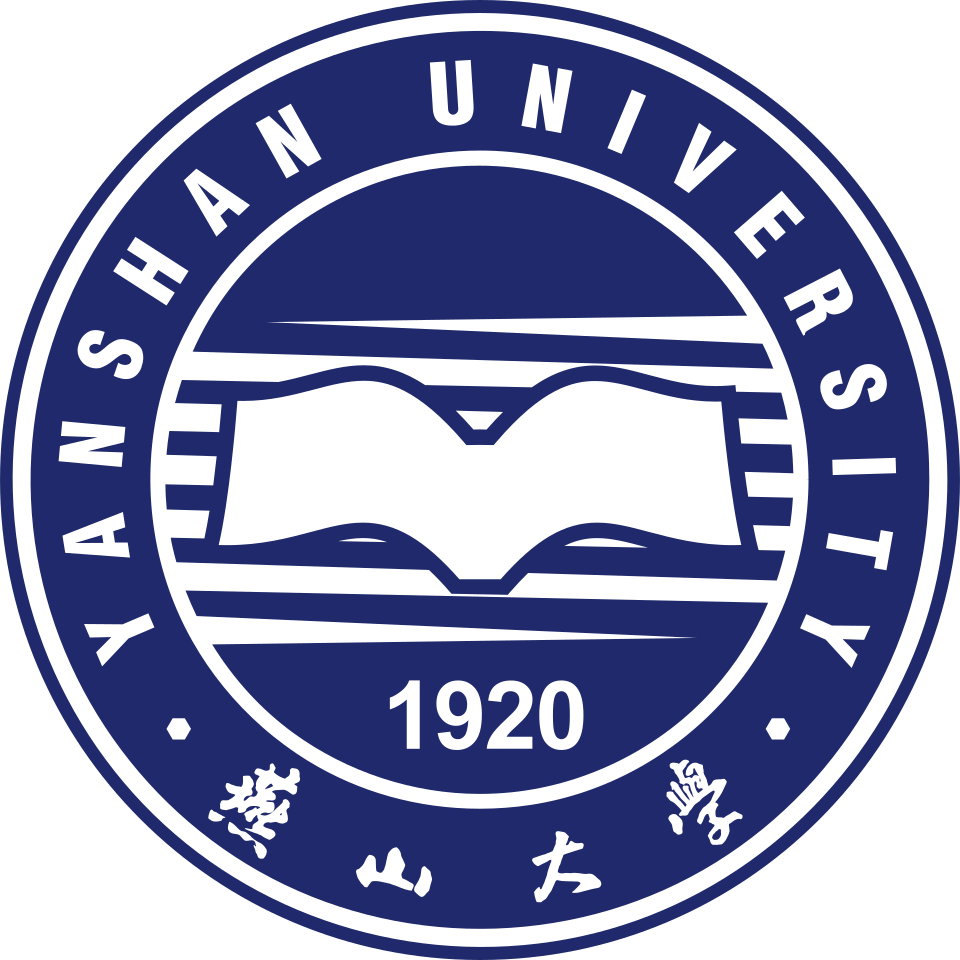}}

\begin{document}

\begin{abstract}
Agents built on large language models (LLMs) build and run software through tool calls.
A call reaches its program through several hops, and any hop can change the call without notice.
When the changed call fails, the agent retries a correct call, which costs users time and money.
Benchmarks and failure analyses do not see the change, because they read the call and its result but not what a hop received.
We define \emph{intent-execution correspondence} (IEC) as the property that the executed action matches the action the emitted call denotes under the tool contract.
Our protocol observes what each hop received without executing the call, and names the first hop that changed it by the receiver's own parser.
\system then delivers the call in a form that this hop cannot alter, or refuses the call.
We build \iecbench from the changes observed in real-world use, with chains of dependent calls under the execution paths of 4 widely-used harnesses.

In 47{,}828 shell calls within production sessions, Claude Code's Bash tool changes 12.0\% of the calls that carry code, escape sequences, or long text.
For 80.7\% of the calls whose backslashes are changed, the wrong action runs without any reported error.
All 10 measured harnesses change a call.
Trajectory-based judgment attributes 95.1\% of the production failures to the LLM, although the path caused more than half of them.
On \iecbench, the path raises the token cost per passed task 2.4 times (up to 12.3 times).
A hop that changes a call also hides the changes after it, so 55.1\% of the failures on one path appear only after its first hop is repaired.
\system, deployed in a commercial product, recovers 79.2\% of the failures with a changed call.
Harnesses should therefore be designed and tested hop-by-hop to ensure a correct call executes as intended or is refused.
\end{abstract}

\maketitle

\section{Introduction}
\label{sec:introduction}

Agents built on large language models (LLMs) build and run software through tool calls~\citep{yao2023react,liu2026agentsurvey}.
A tool call reaches its program over an execution path: the harness serializes the arguments, a host wrapper passes them to an interpreter, the operating system forwards them, and the program parses them again.\footnote{\url{https://learn.microsoft.com/en-us/powershell/module/microsoft.powershell.core/about/about_parsing}}
Each of these hops can change the call without notice~\citep{li2026quotebench}, as the 2 recorded runs of Figure~\ref{fig:path} show.
The program then executes an action that the agent did not intend.
Problems of command execution account for 25.0\% of 3{,}864 bug reports of 3 coding agents~\citep{zhang2026engineering}.
They cost users failed runs and retries, and therefore time and money~\citep{roychoudhury2026trust,watanabe2026agentic}.
However, benchmarks score the selected call~\citep{patil2025bfcl,ye2025toolhop} or the final state~\citep{jimenez2024swebench} and assume that the call is executed as written.
A failure that the path caused is therefore incorrectly attributed to the LLM, and these scores decide which LLM is adopted and which harness is deployed~\citep{xia2025demystifying}.
Failure analyses do not correct the attribution, because the trajectory records the call and its result but not what a hop received~\citep{zhang2025whowhen}.

\definecolor{f1nBlue}{RGB}{33,102,172}
\definecolor{f1nRed}{RGB}{190,30,45}
\definecolor{f1nGreen}{RGB}{35,120,60}
\definecolor{f1nGray}{RGB}{110,110,110}
\definecolor{f1nInk}{RGB}{45,45,45}

\begin{figure}[htbp]
\centering
\hyphenpenalty=10000 \exhyphenpenalty=10000
\def\fnS{\fontsize{8}{9.6}\selectfont}%
\def\fnF{\fontsize{7}{8.4}\selectfont}%
\def\fnB{\fontsize{12}{12}\selectfont}%
\def\fnI{\fontsize{16}{16}\selectfont}%
\begin{tikzpicture}[x=1pt, y=1pt,
  font=\rmfamily\fnF, line width=0.5pt,
  panel/.style n args={3}{draw=#1!55, fill=#1!#2, rounded corners=3pt, line width=0.5pt, minimum width=#3, minimum height=68pt, anchor=north west, inner sep=0pt, outer sep=0pt},
  lab/.style={anchor=base west, inner sep=0pt, outer sep=0pt, font=\rmfamily\fnS\bfseries, text=#1},
  note/.style={anchor=base east, inner sep=0pt, outer sep=0pt, font=\rmfamily\fnF, text=#1},
  code/.style={anchor=base west, inner sep=0pt, outer sep=0pt, font=\ttfamily\fnF, text=f1nInk},
  codeb/.style={code, font=\ttfamily\fnF\bfseries, text=f1nRed},
  codee/.style={code, text=f1nRed!85!black},
  say/.style={anchor=base west, inner sep=0pt, outer sep=0pt, font=\rmfamily\fnF, text=#1},
  sayc/.style={anchor=base, inner sep=0pt, outer sep=0pt, font=\rmfamily\fnF\bfseries, text=#1},
  hdr/.style={anchor=base west, inner sep=0pt, outer sep=0pt, font=\rmfamily\fnS\bfseries, text=f1nInk},
  bigicon/.style={anchor=center, inner sep=0pt, outer sep=0pt, font=\fnI, text=#1},
  chip/.style={anchor=base west, draw=#1, fill=white, text=#1, rounded corners=2pt, inner xsep=1.5pt, inner ysep=2.2pt, text height=5pt, text depth=1.5pt, font=\ttfamily\fnF\bfseries, line width=0.5pt, outer sep=0pt},
  bubble/.style={anchor=north west, draw=f1nGray!60, fill=white, rounded corners=2.5pt, inner xsep=3.5pt, inner ysep=2.6pt, align=left, font=\rmfamily\fnF, text=f1nInk, line width=0.5pt, outer sep=0pt},
  pill/.style={anchor=north east, draw=f1nGreen!55, fill=f1nGreen!8, rounded corners=5.25pt, inner xsep=5.5pt, inner ysep=2pt, text height=5pt, text depth=1.5pt, font=\rmfamily\fnF, text=f1nGreen!80!black, line width=0.5pt, outer sep=0pt},
  rec/.style={anchor=north east, draw=f1nGray!50, fill=f1nGray!10, rounded corners=5.25pt, inner xsep=5pt, inner ysep=2pt, text height=5pt, text depth=1.5pt, font=\rmfamily\fnF, text=f1nGray!85!black, line width=0.5pt, outer sep=0pt},
  tile/.style 2 args={draw=#1!45, fill=#1!5, rounded corners=3pt, minimum height=21pt, minimum width=#2, anchor=north west, inner sep=0pt, outer sep=0pt, line width=0.5pt},
  big/.style={anchor=base west, inner sep=0pt, outer sep=0pt, font=\rmfamily\fnB\bfseries, text=#1},
  cut/.style={f1nRed!80, dash pattern=on 1.6pt off 1.2pt, line width=0.6pt},
]
\def\pad{3.5}
\newcommand{\chev}[3]{\fill[#3] (#1-1.75,#2+2.5) -- (#1+1.75,#2) -- (#1-1.75,#2-2.5) -- cycle;}
\newcommand{\tail}[2]{\fill[white, draw=f1nGray!60, line width=0.5pt] ($(#1.north)+(#2-2.5,0)$) -- ($(#1.north)+(#2,3.2)$) -- ($(#1.north)+(#2+2.5,0)$);
  \draw[white, line width=0.9pt] ($(#1.north)+(#2-2.2,0)$) -- ($(#1.north)+(#2+2.2,0)$);}

\node[hdr] at (0.25,-7) {(a)\ \ One call, changed without a trace};
\def\yA{-13}
\node[panel={f1nBlue}{7}{91pt}]   (a1) at (0.25,\yA) {};
\node[panel={f1nRed}{8}{74pt}]    (a2) at (97.75,\yA) {};
\node[panel={f1nRed}{5}{92pt}]    (a3) at (178.25,\yA) {};
\node[panel={f1nGray}{10}{114pt}] (a4) at (276.75,\yA) {};
\chev{94.5}{\yA-34}{f1nBlue}\chev{175}{\yA-34}{f1nRed}\chev{273.5}{\yA-34}{f1nRed}
\node[lab=f1nBlue] at ($(a1.north west)+(\pad,-10.5)$) {\faIcon{robot}\ \ Intended};
\node[code] at ($(a1.north west)+(\pad,-20.5)$) {python -c "};
\draw[cut] ($(a1.north west)+(\pad,-24.5)$) -- ($(a1.north west)+(\pad+56,-24.5)$);
\node[anchor=west, inner sep=0pt, outer sep=0pt, text=f1nRed, font=\fnF] at ($(a1.north west)+(\pad+58,-24.5)$) {\faIcon{cut}};
\foreach \w [count=\i from 0] in {28,20,32,24,16,28,22,30,18,26,32,20,24,16,28,22,18,10}
  \draw[f1nGray!50, line width=0.8pt] ($(a1.north west)+(\pad+1.5,-29.5-1.75*\i)$) -- ++(\w,0);
\node[anchor=west, inner sep=0pt, outer sep=0pt, text=f1nGray, font=\rmfamily\fnF\itshape] at ($(a1.north west)+(\pad+38,-44)$) {18-line program};
\node[lab=f1nRed] at ($(a2.north west)+(\pad,-10.5)$) {hop: shell parse};
\node[bigicon=f1nRed] at ($(a2.north west)+(37,-22.5)$) {\faIcon{cut}};
\node[codeb] at ($(a2.north west)+(\pad,-38.5)$) {cmd.exe /d /s /c};
\node[say=f1nRed!85!black] at ($(a2.north west)+(\pad,-47.5)$) {ends the command at};
\node[say=f1nRed!85!black] at ($(a2.north west)+(\pad,-56)$) {its first line break};
\node[lab=f1nRed] at ($(a3.north west)+(\pad,-10.5)$) {\faIcon{microchip}\ \ Executed};
\node[code] at ($(a3.north west)+(\pad,-20.5)$) {python -c "};
\draw[cut] ($(a3.north west)+(\pad,-24.5)$) -- ($(a3.north west)+(\pad+56,-24.5)$);
\draw[f1nRed!40, dash pattern=on 0.6pt off 1.2pt, line width=0.5pt, rounded corners=2pt] ($(a3.north west)+(\pad,-28.5)$) rectangle ($(a3.north west)+(92-\pad,-46.5)$);
\node[sayc=f1nRed] at ($(a3.north west)+(46,-39.7)$) {an empty program};
\node[code] at ($(a3.north west)+(\pad,-54)$) {Output: (empty)};
\node[code] at ($(a3.north west)+(\pad,-62.5)$) {Exit Code: 0};
\node[lab=f1nGray] at ($(a4.north west)+(\pad,-10.5)$) {\faIcon{eye}\ \ Agent};
\node[bubble] (ab) at ($(a4.north west)+(\pad,-16.5)$) {{\itshape``The comma decimal separator}\\{\itshape might be causing issues \dots\ Let me}\\{\itshape use pandas.''}\\reports done $\rightarrow$ {\bfseries\color{f1nRed}reward 0}};
\tail{ab}{12}
\node[say=f1nGray] at (0.25,\yA-77.5) {Terminal-Bench~2 run\ $\cdot$\ Qwen Code on Windows};
\node[pill] at (390.75,\yA-70.5) {{\color{f1nGreen}\faIcon{check-circle}}\ \ \textbf{Control}\ \ on Linux, the program runs whole};

\node[hdr] at (0.25,-104) {(b)\ \ One value, changed in the middle of a chain};
\def\yB{-110}
\node[panel={f1nBlue}{7}{91pt}]   (b1) at (0.25,\yB) {};
\node[panel={f1nRed}{8}{74pt}]    (b2) at (97.75,\yB) {};
\node[panel={f1nRed}{5}{92pt}]    (b3) at (178.25,\yB) {};
\node[panel={f1nGray}{10}{114pt}] (b4) at (276.75,\yB) {};
\chev{94.5}{\yB-34}{f1nBlue}\chev{175}{\yB-34}{f1nRed}\chev{273.5}{\yB-34}{f1nRed}
\node[lab=f1nBlue] at ($(b1.north west)+(\pad,-10.5)$) {\faIcon{robot}\ \ Intended};
\node[code] at ($(b1.north west)+(\pad,-20.5)$) {genealogy\_search};
\node[code] at ($(b1.north west)+(\pad,-29)$) {-{}-person\_name};
\node[chip=f1nBlue] at ($(b1.north west)+(\pad,-41.5)$) {Joseph-\'Edouard Perrault};
\node[code] at ($(b1.north west)+(\pad,-54)$) {-{}-relationship father};
\node[lab=f1nRed] at ($(b2.north west)+(\pad,-10.5)$) {hop: host wrapper};
\node[bigicon=f1nRed] at ($(b2.north west)+(37,-22.5)$) {\faIcon{quote-right}};
\node[codeb] at ($(b2.north west)+(\pad,-37.5)$) {powershell.exe};
\node[codeb] at ($(b2.north west)+(\pad,-46)$) {-Command {\normalfont\rmfamily\itshape\fnF text}};
\node[say=f1nRed!85!black] at ($(b2.north west)+(\pad,-54.5)$) {quotes consumed};
\node[say=f1nRed!85!black] at ($(b2.north west)+(\pad,-63)$) {as delimiters};
\node[lab=f1nRed] at ($(b3.north west)+(\pad,-10.5)$) {\faIcon{microchip}\ \ Executed};
\node[code] at ($(b3.north west)+(\pad,-20.5)$) {genealogy\_search};
\node[code] at ($(b3.north west)+(\pad,-29)$) {-{}-person\_name};
\node[chip=f1nRed] at ($(b3.north west)+(\pad,-41.5)$) {Joseph-\'Edouard};
\node[chip=f1nRed] at ($(b3.north west)+(\pad+54,-41.5)$) {Perrault};
\node[codee] at ($(b3.north west)+(\pad,-54)$) {error: unrecognized};
\node[codee] at ($(b3.north west)+(\pad,-62.5)$) {arguments: Perrault};
\node[lab=f1nGray] at ($(b4.north west)+(\pad,-10.5)$) {\faIcon{eye}\ \ Agent};
\node[bubble] (bb) at ($(b4.north west)+(\pad,-19)$) {re-sends {\ttfamily -{}-person\_name}\\{\ttfamily "Joseph-\'Edouard"}\\drops the surname,\\answers {\bfseries\color{f1nRed}1849-09-04}};
\tail{bb}{12}
\node[say=f1nGray] at (0.25,\yB-77.5) {\textsc{IEC-Bench} chain\ $\cdot$\ correct answer 1846-05-18};
\node[pill] at (390.75,\yB-70.5) {{\color{f1nGreen}\faIcon{check-circle}}\ \ \textbf{Control}\ \ command from a file: \textbf{1846-05-18}};

\end{tikzpicture}
\caption{A hop changes a correct call: silently (a), with a misleading error (b).}
\label{fig:path}
\end{figure}
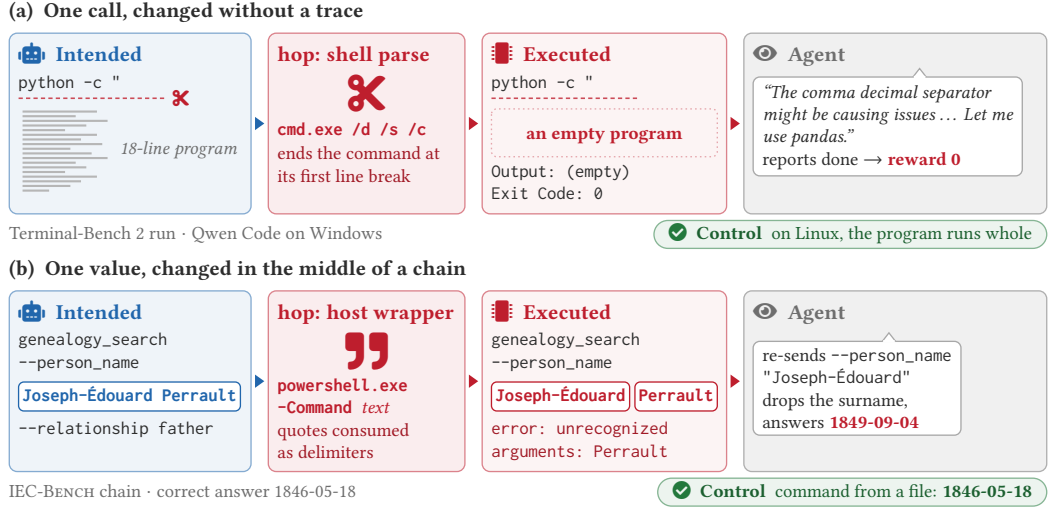

Existing research on agent reliability mostly examines what the LLM did.
Who\&When~\citep{zhang2025whowhen} names the responsible agent and step from the trajectory.
Empirical studies classify failures from trajectories~\citep{liu2025failures,majgaonkar2025behaviour,bouzenia2025trajectories}.
Other techniques act on the agent rather than on the path.
Fault injection tests an agent system~\citep{tan2026agentchaos,jia2026masfire,seyedghorban2026observability}.
Causal Agent Replay~\citep{shah2026car} re-executes a run from one of its steps.
Reflexion~\citep{shinn2023reflexion} lets the agent retry.
ToolBench-X~\citep{tian2026toolbenchx} injects visible tool-environment hazards, and Gupta et al.~\citep{gupta2026dissecting} return structured execution feedback.
Further studies examine defects between LLM and harness~\citep{chen2026arbugs,ning2026defects}.
Only a few studies examine the path itself.
QuoteBench~\citep{li2026quotebench} measures the wrapper, and BatBadBut documents batch-file argument passing.\footnote{\url{https://flatt.tech/research/posts/batbadbut-you-cant-securely-execute-commands-on-windows/}}
None of these studies establishes whether the executed action matches the action the emitted call denotes under the tool contract.
We call this property \emph{intent-execution correspondence} (IEC).

\emph{Three gaps} stand in the way of measuring and repairing IEC:

\begin{itemize}[leftmargin=*,topsep=2pt,itemsep=1pt]
\item[\ding{172}] \textbf{Trajectory-based analyses do not observe the path.}
Who\&When~\citep{zhang2025whowhen} reads the trajectory of a failed run.
The trajectory does not contain what a hop received, so a changed call is incorrectly attributed to the LLM.
\item[\ding{173}] \textbf{Studies of the path examine 1 hop.}
QuoteBench~\citep{li2026quotebench} records the wrapper on 56 one-shot Bash tasks against a single reference path.
However, an agent's path has 5 hops (Figure~\ref{fig:method}), and a hop that changes a call hides the changes after it.
The hop that changed a call also determines the repair.
\item[\ding{174}] \textbf{Real-world use is unmeasured, and interventions target the agent or one hop.}
QuoteBench and ToolBench-X~\citep{tian2026toolbenchx} run benchmark tasks, and Zhang et al.~\citep{zhang2026engineering} analyze bug reports, so none shows how often and at which hop calls change in real-world use.
Causal Agent Replay~\citep{shah2026car}, Reflexion~\citep{shinn2023reflexion}, Gupta et al.~\citep{gupta2026dissecting}, and QuoteBench's temporary file act on a step, on the feedback, or on the wrapper alone.
Validating an attribution, however, requires repairing that hop alone.
\end{itemize}

To address these gaps, we propose the IEC protocol, which measures the path, and \system, which repairs it.
A recorded call cannot be executed again safely, so the IEC protocol observes what each hop received without executing the call (\ding{172}).
Different texts can denote the same action, so the receiver's own parser decides whether the action changed, and the IEC protocol names the first hop that changed it (\ding{173}).
A hop that changes a call hides the changes after it, so the IEC protocol confirms an attribution by re-executing the recorded run with only that hop repaired (\ding{174}).
\system then delivers the call in a form that this hop cannot alter, or refuses the call.
The hop determines the form, so one technique repairs a path at its wrapper, at the hand-over to the program, or at both.

The main contributions of this work are as follows:
\begin{itemize}[noitemsep,topsep=0pt,leftmargin=2em]
\item \textbf{Problem and finding.}
We define intent-execution correspondence for tool calls and chains of dependent calls, and find that a hop that changes a call hides the changes after it.
On a path with 2 such hops, 38 of the 69 failures that the path caused appear only after the first hop is repaired.
\item \textbf{Measurement of real-world use.}
The first measurement of changed calls in real-world use covers 47{,}828 shell calls of 261 production sessions, 9 developers on Linux, 10 harnesses, 6 shells, 22 MCP servers, and 3 public trajectory sources.
Claude Code's Bash tool changes 12.0\% of the calls that carry code, escape sequences, or long text.
For 80.7\% of the calls whose backslashes are changed, the wrong action runs without any reported error.
\item \textbf{Evaluation method.}
The IEC protocol observes each hop without executing the call and, by the receiver's own parser, names the first hop that changed the call.
We validate the IEC protocol by mutation testing and fault injection.
We build \iecbench from changes observed in real-world use, with 67 task pairs, 135 chains, and 46 Terminal-Bench~2 tasks under 6 execution paths.
\item \textbf{Solution and validation.}
\system repairs the path at the hop that the IEC protocol names, composes repairs where an earlier hop hides a later one, and refuses a call that it cannot deliver as written.
It recovers 79.2\% of the \iecbench failures with a changed call and lowers the token cost per passed task 2.4 times.
We deploy it in production and in a commercial product, and release it as a Claude Code mod, a shell shim, and an MCP server.
We compare the IEC protocol and \system with published attribution and repair baselines on 4 LLMs.
\end{itemize}

\section{Approach}
\label{sec:approach}

\subsection{Intent-Execution Correspondence}
\label{sec:model}

The intended action $a$ of a tool call is the action the emitted call denotes under the tool contract, that is, the shell or argument semantics that the tool description states to the LLM.
We take the emitted call as the intent, not the LLM's or the user's goal~\citep{feng2025tai3}, so that a change by the path can be told from a generation error.
A call whose text is incorrect for its shell is an incorrect intent executed as written, and its failure is a generation error.
For a command, the action is its executable, ordered argument vector, control operators, working directory, task-relevant environment, and standard input.
For a structured tool, the action is its bound argument values.
A \emph{harness} is the program that runs the agent, and a \emph{launch} is the way a harness starts a shell or a program.
The call reaches the program over the execution path, a sequence of \emph{hops} $\Pi=\langle b_1,\ldots,b_m\rangle$.
At each hop, one side serializes the call and the receiving side parses it.
Let $r_i$ be the action that the receiver of hop $b_i$ parses, in the working directory and environment of the launch.
The path \emph{corresponds} for $a$ when every hop receives the intended action, $r_i = a$ for all $i$.
The first hop that receives a different action is the \emph{first divergence}, $\mathsf{FD}(a,\Pi)=\min\{i \mid r_i \neq a\}$.
Section~\ref{sec:fd} states how the receiver's own parser decides whether $r_i = a$.
A divergence is \emph{silent} when the tool result returned to the agent contains neither an error flag nor a parser or quoting error message, and \emph{loud} otherwise.

In a chain of calls, a change at one call propagates to the calls after it.
A \emph{step} is one call of a chain $a_1,\ldots,a_n$, in which the arguments of $a_j$ may be values returned by earlier calls, whereas a hop lies on the path of one call.
The chain corresponds when every call in it corresponds.
Later calls use the result of a changed call, so the chain can end in an incorrect answer even if every later call arrives as written.
A loud divergence makes the agent revise its intent after an error that refers to its own text.

A hop changes a call through 4 mechanisms.
In \emph{quote consumption}, text appended to a command line is parsed a second time, so \code{console.log("test")} arrives as \code{console.log(test)}.
In \emph{re-expansion}, a receiver reinterprets a delivered literal, \code{\%PATH\%} in a batch operand or \code{!x} in an interactive Bash.
In \emph{truncation}, the MSYS2 runtime of Git Bash cuts a quoted argument at 8{,}186 UTF-16 code units\footnote{\url{https://github.com/zetaloop/msys2-argv-fix}} and collapses its backslash pairs\footnote{\url{https://github.com/msys2/msys2-runtime/issues/36}}, and \code{cmd.exe} ends a command at its first line break.
In an \emph{arity change}, an operand is dropped or split.
Neither stdout, stderr, nor the exit code names the mechanism.
The same mechanism is repaired differently at different hops, so the hop determines the repair.

\definecolor{f2nBlue}{RGB}{33,102,172}
\definecolor{f2nRed}{RGB}{190,30,45}
\definecolor{f2nPur}{RGB}{120,80,150}
\definecolor{f2nGreen}{RGB}{35,125,70}
\definecolor{f2nBlueBg}{RGB}{236,244,252}
\definecolor{f2nRedBg}{RGB}{253,237,237}
\definecolor{f2nPurBg}{RGB}{245,240,250}
\definecolor{f2nGreenBg}{RGB}{235,246,238}
\definecolor{f2nGray}{RGB}{110,110,110}
\definecolor{f2nGrayBg}{RGB}{243,243,243}
\definecolor{f2nLaneBg}{RGB}{247,249,253}
\definecolor{f2nInk}{RGB}{70,70,70}
\definecolor{f2nLane}{RGB}{240,242,247}
\providecommand{\fmHop}[6]{%
  \pgfmathsetmacro{\fmA}{#1+2.036}\pgfmathsetmacro{\fmB}{#1+2.316}\pgfmathsetmacro{\fmN}{#1+0.28}%
  \pgfmathsetmacro{\fmC}{#1+1.158}\pgfmathsetmacro{\fmD}{#1+0.30}%
  \draw[hopshape] (#1,0) -- (\fmA,0) -- (\fmB,-0.975) -- (\fmA,-1.95) -- (#1,-1.95) -- (\fmN,-0.975) -- cycle;
  \node[badge] at (\fmD,-0.26) {#2};
  \node[hicon] at (\fmC,-0.50) {\faIcon{#3}};
  \node[lbl] at (\fmC,-1.05) {#4};
  \node[lbl] at (\fmC,-1.38) {#5};
  \node[mono] at (\fmC,-1.74) {#6};
}
\providecommand{\fmLet}[2]{\tikz[baseline=(fmc.base)]{\node[circle, draw=#1, line width=0.6pt, inner sep=0.9pt, text=#1, font=\rmfamily\bfseries\fontsize{8.5}{8.5}\selectfont] (fmc) {#2};}}
\providecommand{\fmS}{\fontsize{7.5}{9}\selectfont}

\begin{figure*}[htbp]
\centering
\begin{tikzpicture}[
  font=\rmfamily\fontsize{8}{9.6}\selectfont,
  >={Latex[length=1.4mm,width=1.2mm]},
  lane/.style={fill=f2nLane, rounded corners=4pt},
  lanelbl/.style={anchor=base west, inner sep=0pt, text=f2nGray, font=\rmfamily\bfseries\fontsize{8}{9.6}\selectfont},
  hopshape/.style={draw=black!40, line width=0.5pt, fill=white, rounded corners=1.2pt},
  endshape/.style={line width=0.5pt, rounded corners=1.2pt},
  badge/.style={circle, fill=black!65, text=white, font=\rmfamily\bfseries\fontsize{7}{8.4}\selectfont, inner sep=0pt, minimum size=3.2mm},
  hicon/.style={inner sep=0pt, text=black!60, font=\LARGE},
  lbl/.style={anchor=base, inner sep=0pt, text=black!85},
  mono/.style={anchor=base, inner sep=0pt, text=f2nInk},
  tx/.style={anchor=base west, inner sep=0pt, text=black!80},
  mod/.style={draw=#1, line width=0.5pt, rounded corners=3pt},
  chip/.style={draw=f2nRed!55, line width=0.5pt, fill=white, rounded corners=2pt, inner xsep=3pt, inner ysep=0pt, minimum height=3.6mm, text width=2.60cm, align=left, text=black!85, anchor=north west, outer sep=0pt, font=\rmfamily\fmS},
  flowg/.style={->, line width=0.8pt, black!55},
  flowb/.style={->, line width=1.0pt, f2nBlue!85, rounded corners=2pt},
  flowgr/.style={->, line width=1.0pt, f2nGreen!85, line join=round},
  modstep/.style={->, line width=0.8pt, black!50},
  tap/.style={f2nPur!80, line width=0.7pt},
  fmrun/.style={->, line width=0.8pt, f2nPur!80, rounded corners=2pt},
  fmpill/.style={draw=#1!65, line width=0.6pt, fill=white, rounded corners=1.9mm, inner sep=0pt, minimum height=3.8mm, minimum width=1.85cm, anchor=west, text=#1!85!black},
  fmmark/.style={inner sep=0pt, font=\large},
  fmibox/.style={draw=f2nGreen!65, line width=0.6pt, fill=white, rounded corners=2pt, inner sep=0pt, minimum height=3.9mm, text=black!85, font=\rmfamily\fmS},
  fmtt/.style={line width=0.5pt, fill=white, rounded corners=2pt, inner xsep=3pt, inner ysep=2pt, text=black!85, font=\ttfamily\fontsize{7}{8.4}\selectfont},
  fmstage/.style={draw=f2nGreen!65, line width=0.6pt, fill=white, rounded corners=2pt, inner sep=2pt, align=center, anchor=west, minimum height=7.2mm, text=black!85, font=\rmfamily\fmS},
  fmdone/.style={draw=f2nGreen!85, line width=0.9pt, fill=f2nGreenBg},
  fmtaken/.style={->, line width=1.0pt, f2nGreen!85},
  fmrow/.style={anchor=west, inner sep=0pt, text=black!80, font=\rmfamily\fmS},
  fmcell/.style={draw=#1!65, line width=0.6pt, fill=white, rounded corners=1.5pt, inner sep=0pt, minimum size=3.8mm, font=\rmfamily\fmS},
  fmkey/.style={anchor=base west, inner sep=0pt, text=f2nGray, font=\ttfamily\fmS},
  fmval/.style={anchor=base west, inner sep=0pt, text=black!85, font=\ttfamily\fmS},
  fmarg/.style={anchor=base west, draw=f2nRed!55, line width=0.5pt, fill=white, rounded corners=2pt, inner xsep=3pt, inner ysep=2pt, text=f2nRed!85!black, font=\ttfamily\fmS},
]
\fill[rounded corners=4pt, fill=f2nGreenBg] (-0.06,4.06) rectangle (13.64,0.60);
\fill[lane] (-0.06,0.50) rectangle (13.64,-2.02);
\fill[lane] (-0.06,-2.12) rectangle (13.64,-4.51);
\node[lanelbl, anchor=base east] at (13.54,0.20) {the path};
\node[lanelbl] at (0.06,-2.36) {the IEC protocol};
\draw[endshape, draw=f2nBlue!60, fill=f2nBlueBg] (0,0) -- (1.02,0) -- (1.30,-0.975) -- (1.02,-1.95) -- (0,-1.95) -- cycle;
\node[inner sep=0pt, text=f2nBlue, font=\LARGE] at (0.58,-0.70) {\faIcon{robot}};
\node[lbl] at (0.58,-1.30) {\textbf{agent}};
\fmHop{1.100}{1}{exchange-alt}{tool}{serialization}{\texttt{JSON}\,$\to$\,\texttt{text}}
\fmHop{3.216}{2}{cogs}{host}{wrapper}{\texttt{eval '...'}}
\fmHop{5.332}{3}{terminal}{shell}{parse}{\texttt{Bash}\,$\cdot$\,\texttt{PS}\,$\cdot$\,\texttt{cmd}}
\fmHop{7.448}{4}{microchip}{process}{interface}{\texttt{line}\,$\to$\,\texttt{argv}}
\fmHop{9.564}{5}{file-code}{target}{parser}{\texttt{its argv}}
\draw[endshape, draw=black!50, fill=f2nGrayBg] (11.68,0) -- (13.28,0) -- (13.28,-1.95) -- (11.68,-1.95) -- (11.96,-0.975) -- cycle;
\node[inner sep=0pt, text=black!70, font=\LARGE] at (12.60,-0.70) {\faIcon{play-circle}};
\node[lbl] at (12.60,-1.30) {\textbf{program}};
\draw[flowg, rounded corners=2pt] (12.48,0) -- (12.48,0.17) -- (0.78,0.17) -- (0.78,0);
\node[inner sep=1pt, fill=f2nLane, text=f2nGray, font=\rmfamily\fmS] at (6.49,0.17) {tool result};
\draw[tap] (2.258,-2.55) -- (2.258,-2.28) -- (10.722,-2.28);
\foreach \x in {2.258,4.374,6.490,8.606,10.722} {\draw[tap] (\x,-2.28) -- (\x,-1.95); \fill[f2nPur!80] (\x,-1.95) circle (0.55mm);}
\draw[mod=f2nPur!60, fill=f2nPurBg] (0,-2.550) rectangle (3.40,-4.390);
\node[tx] at (0.106,-3.030) {\fmLet{f2nPur}{A}~\textcolor{f2nPur}{\Large\faIcon{eye}}~\textcolor{f2nPur!85!black}{\textbf{witness}}};
\node[tx] at (0.106,-3.390) {arrived text, not executed};
\node[tx, text=f2nPur!85!black, font=\rmfamily\fmS] at (0.106,-3.770) {\textcolor{black!65}{\ding{182}}\,recorded call\ \textcolor{black!65}{\ding{183}\ding{184}}\,\texttt{BASH\_ENV}};
\node[tx, text=f2nPur!85!black, font=\rmfamily\fmS] at (0.106,-4.130) {\textcolor{black!65}{\ding{185}}\,argv probe\enspace\textcolor{black!65}{\ding{186}}\,argv sentry};
\draw[mod=f2nRed!60, fill=f2nRedBg] (3.64,-2.550) rectangle (8.90,-4.390);
\node[tx] at (3.746,-3.030) {\fmLet{f2nRed}{B}~\textcolor{f2nRed}{\Large\faIcon{balance-scale}}~\textcolor{f2nRed}{\textbf{first divergence}}};
\node[tx] at (3.746,-3.370) {the receiver's own parse};
\node[chip] at (3.746,-3.520) {\textcolor{f2nRed}{\faIcon{quote-right}}\,quote consumption};
\node[chip, text width=1.95cm] at (6.63,-3.520) {\textcolor{f2nRed}{\faIcon{random}}\,re-expansion};
\node[chip] at (3.746,-3.930) {\textcolor{f2nRed}{\faIcon{cut}}\,truncation};
\node[chip, text width=1.95cm] at (6.63,-3.930) {\textcolor{f2nRed}{\faIcon{list-ol}}\,arity change};
\draw[mod=f2nPur!60, fill=f2nPurBg] (9.14,-2.550) rectangle (13.28,-4.390);
\node[tx] at (9.246,-3.030) {\fmLet{f2nPur}{C}~\textcolor{f2nPur}{\Large\faIcon{save}}~\textcolor{f2nPur!85!black}{\textbf{repair replay}}};
\node[fmpill=f2nRed, minimum width=1.62cm] at (9.25,-3.520) {original path};
\node[fmpill=f2nGreen, minimum width=1.62cm] at (9.25,-3.980) {hop repaired};
\draw[fmrun, f2nRed!80] (10.87,-3.520) -- (11.10,-3.520);
\draw[fmrun, f2nGreen!85] (10.87,-3.980) -- (11.10,-3.980);
\node[fmmark, text=f2nRed] at (11.30,-3.520) {\faIcon{times-circle}};
\node[fmmark, text=f2nGreen] at (11.30,-3.980) {\faIcon{check-circle}};
\draw[black!50, line width=0.6pt, rounded corners=1pt] (11.52,-3.520) -- (11.62,-3.520) -- (11.62,-3.980) -- (11.52,-3.980);
\draw[->, black!50, line width=0.6pt] (11.62,-3.750) -- (11.84,-3.750);
\node[tx, anchor=west, font=\rmfamily\fmS] at (11.88,-3.750) {path-caused};
\node[tx] at (0.106,3.64) {\textcolor{f2nGreen}{\Large\faIcon{wrench}}~\textcolor{f2nGreen!85!black}{\textbf{\textsc{IntAct}: each call is rendered for the named hops before the launch, or refused}}};
\draw[draw=f2nGreen!70, line width=0.8pt, fill=white, rounded corners=3pt] (0.02,0.72) rectangle (13.30,1.66);
\foreach \x/\t in {4.374/{\ding{183} receives}, 6.490/{\ding{184} receives}, 8.606/{\ding{185} receives}, 11.30/{the program receives}} {\node[inner sep=0pt, text=f2nGray, font=\rmfamily\fmS, anchor=base] at (\x,3.16) {\t};}
\node[fmrow, text=f2nGray] at (0.12,3.16) {one call, \texttt{print(\textquotedbl hi\textquotedbl)}};
\node[fmrow] at (0.12,2.82) {original launch};
\node[fmrow] at (0.12,2.36) {only \textcolor{black!65}{\ding{183}} rendered};
\node[fmrow, text=f2nGreen!55!black] at (0.12,1.42) {\textcolor{f2nGreen}{\faIcon{wrench}}\,in the domain: \textcolor{black!65}{\ding{183}} and \textcolor{black!65}{\ding{185}}};
\node[fmrow, text=f2nGreen!55!black] at (0.12,0.96) {\textcolor{f2nGreen}{\faIcon{wrench}}\,outside the domain};
\node[fmibox, anchor=west, inner xsep=4pt, draw=f2nRed!60, text=f2nRed!85!black] at (3.62,0.96) {\faIcon{ban}\,refused before the launch: nothing is executed, and the tool result names the hop};
\node[fmtt, draw=black!40, ] at (4.374,2.82) {\textquotesingle print(\textquotedbl hi\textquotedbl )\textquotesingle };
\node[fmtt, draw=f2nRed!60, text=f2nRed!85!black, ] at (6.490,2.82) {\textquotesingle print(hi)\textquotesingle };
\node[fmtt, draw=f2nRed!60, text=f2nRed!85!black, ] at (8.606,2.82) {print(hi)};
\node[fmtt, draw=f2nRed!60, text=f2nRed!85!black, ] at (11.30,2.82) {print(hi)};
\node[fmtt, draw=black!40, ] at (4.374,2.36) {\faIcon{lock}\,cAB5AHQA..};
\node[fmtt, draw=black!40, ] at (6.490,2.36) {\textquotesingle print(\textquotedbl hi\textquotedbl )\textquotesingle };
\node[fmtt, draw=black!40, ] at (8.606,2.36) {print(\textquotedbl hi\textquotedbl )};
\node[fmtt, draw=f2nRed!60, text=f2nRed!85!black, ] at (11.30,2.36) {print(hi)};
\node[fmtt, draw=f2nGreen!75, text=f2nGreen!55!black, ] at (4.374,1.42) {\faIcon{lock}\,cAB5AHQA..};
\node[fmtt, draw=f2nGreen!75, text=f2nGreen!55!black, ] at (6.490,1.42) {\textquotesingle print(\textbackslash\textquotedbl hi\textbackslash\textquotedbl )\textquotesingle };
\node[fmtt, draw=f2nGreen!75, text=f2nGreen!55!black, ] at (8.606,1.42) {print(\textbackslash\textquotedbl hi\textbackslash\textquotedbl )};
\node[fmtt, draw=f2nGreen!75, text=f2nGreen!55!black, ] at (11.30,1.42) {print(\textquotedbl hi\textquotedbl )};
\node[fmmark, text=f2nRed] at (12.45,2.82) {\faIcon{times-circle}};
\node[fmmark, text=f2nRed] at (12.45,2.36) {\faIcon{times-circle}};
\node[fmmark, text=f2nGreen] at (12.45,1.42) {\faIcon{check-circle}};
\node[fmrow] at (0.12,1.90) {only \textcolor{black!65}{\ding{185}} rendered};
\node[fmtt, draw=black!40, ] at (4.374,1.90) {\textquotesingle print(\textbackslash\textquotedbl hi\textbackslash\textquotedbl )\textquotesingle };
\node[fmtt, draw=black!40, ] at (6.490,1.90) {\textquotesingle print(\textquotedbl hi\textquotedbl )\textquotesingle };
\node[fmtt, draw=black!40, ] at (8.606,1.90) {print(\textquotedbl hi\textquotedbl )};
\node[fmtt, draw=f2nRed!60, text=f2nRed!85!black, ] at (11.30,1.90) {print(hi)};
\node[fmmark, text=f2nRed] at (12.45,1.90) {\faIcon{times-circle}};
\draw[f2nGreen!85, line width=1.0pt] (3.176,0.72) -- (3.176,0.02);
\node[circle, fill=f2nGreen!85, text=white, inner sep=1pt, font=\fontsize{6}{7}\selectfont] at (3.176,0.36) {\faIcon{wrench}};
\node[inner sep=1pt, fill=f2nLane, text=f2nGreen!55!black, font=\rmfamily\fmS, anchor=west] at (3.40,0.36) {\textsc{IntAct}: in the domain?};
\end{tikzpicture}
\caption{The IEC protocol and \system.}
\label{fig:method}
\end{figure*}
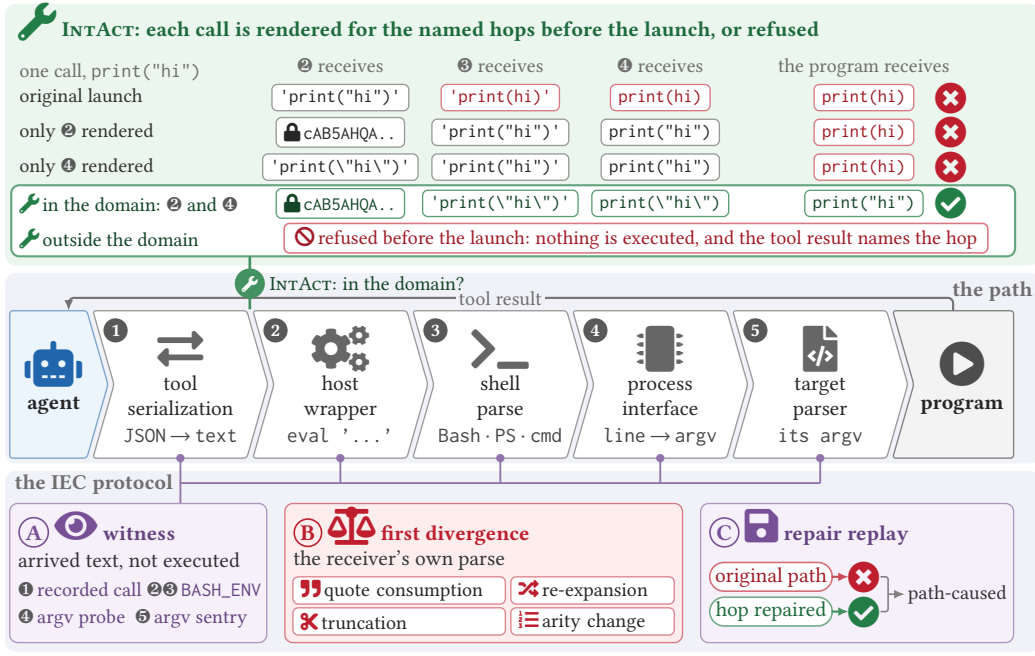

\label{sec:method}Figure~\ref{fig:method} presents \system and the IEC protocol.
The path of a command has 5 hops: the tool serialization, the host wrapper, the shell parse, the process interface, and the target parser.
\system takes each call before the host wrapper and refuses a call that it cannot deliver as written.
Otherwise, it applies the \emph{rendering} of each hop that the IEC protocol names, a form of the call that this hop cannot alter, and passes the rendered text into the launch.
In the IEC protocol, \emph{witnesses} report what each hop received without executing the call (A, Section~\ref{sec:witness}).
The receiver's own parser names the first divergence (B, Section~\ref{sec:fd}).
The \emph{repair replay}, the replay of a recorded run with only that hop repaired, confirms the losses that the path caused (C, Section~\ref{app:dataset}).
For this replay, we record calls and tool results as emitted and received, and a reply cache replays a run without an LLM.

\subsection{Witnesses: Observing a Hop Without Executing the Call}
\label{sec:witness}

A witness reports what a hop received and executes nothing.
Executing a recorded call again is neither safe nor repeatable, and the trajectory holds only the text that the agent sent and the result that came back.
A witness therefore keeps the harness's launch and replaces only the program that would execute the command with one that reports the received text.
For Bash, the witness is the file named by \code{BASH\_ENV}, which Bash sources before it executes the command string, so the reported string has passed through the wrapper as in the harness.
For a launch that appends the command after \code{-Command}, the \emph{appended launch}, the witness computes the script that PowerShell executes by the rules that \code{powershell.exe} applies to its command line.
The call changes at the wrapper when PowerShell's parse of this script differs from its parse of the text.
A \emph{one-argument launch} passes the text to \code{powershell.exe} as one argument, which the wrapper delivers intact.
At the process interface, the witness records the argument vector that a native program receives, which the script does not show.
The witness parses the script without execution and runs each native invocation with constant arguments separately, with an argv probe in place of the program.
With an LLM in the loop, an argv sentry in place of the target program also records the ancestor processes, which gives the wrapper of a running harness.
For structured tools, the witness records the bound values that the tool function receives.

\subsection{Correspondence per Receiver and the First Divergence}
\label{sec:fd}

The IEC protocol decides whether a hop changed a call by the receiver's own parser, not by string equality or judgment.
Texts that differ can denote the same action, as \code{"a\textbackslash\textbackslash b"} and \code{"a\textbackslash b"} denote the same word in Bash, and string comparison classifies 15 of 35 unchanged pairs as changed (Section~\ref{sec:rq3}).
The IEC protocol compares the operators, words, and heredoc bodies of a Bash command, the parse tree of a PowerShell script, and the argument vector of a program.
The comparisons return a verdict for 99.98\% of the 32{,}259 Bash calls of the production sessions.

The IEC protocol attributes a loud failure to generation, to the path, or to a nested hop.
The error message does not tell them apart, because it refers to the agent's text in every case.
The IEC protocol first parses the agent's text with execution disabled, by the interpreter that receives it.
Text that does not parse was incorrect when emitted, a generation error.
Text that parses is delivered again through the witness, and a difference is attributed to the path.
An error at a line beyond the end of the agent's text is also attributed to the path, because that line lies in the text that the wrapper appends.
Text that parses, is received intact, and invokes a further interpreter fails at a nested hop.

\subsection{\texorpdfstring{\textsc{\systemname}}{\systemname}: Repair at the Named Hop}
\label{sec:act}

\system repairs the path at the hop that the IEC protocol names.
The same change needs a different repair at a different hop, and requoting every argument restores 3 of 30 injected faults (Section~\ref{sec:rq4}), so the hop selects the repair.
The IEC protocol names a launch's hops once, when we measure the launch.
The input of \system is the path $\Pi$ of the launch, its named hops $H$, and the text $x$ of the call, and the output is a rendering or a refusal (Table~\ref{tab:intact}, Algorithm~\ref{alg:intact}).
The named hops are the first divergence $\mathsf{FD}(a,\Pi)$ and the second hop of a launch that changes a call twice.

\begin{table}[htbp]
\centering
\caption{Renderings of \system.}
\label{tab:intact}
\footnotesize
\setlength{\tabcolsep}{3pt}
\begin{tabularx}{\columnwidth}{@{}>{\raggedright\arraybackslash}p{0.15\columnwidth}>{\raggedright\arraybackslash}p{0.12\columnwidth}YY>{\raggedright\arraybackslash}p{0.27\columnwidth}@{}}
\toprule
Launch & Named hop & Change at the hop & Rendering & Effect \\
\midrule
Claude Code's Bash launch & host wrapper & argument truncated, backslash pairs collapsed & the wrapper sources a file & all 227 calls rerouted, 20 of 22 lost pairs; 143 $\to$ 0 \\
Qwen Code's \code{cmd.exe} launch & shell parse & command ends at its first line break & multi-line text is refused & 41 $\to$ 0, 54 calls refused \\
One-argument launch & process interface & quotes consumed, arguments split or dropped & native rendering of each literal & 48 of 70 lost pairs; 1{,}094 $\to$ 0 \\
Appended launch & host wrapper, process interface & quotes consumed, arguments split or dropped & \code{-EncodedCommand} and native rendering & 69 of 81 lost pairs; 3{,}919 $\to$ 0 \\
Typing into an interactive Bash & shell parse & history expansion and completion & both switched off before typing & 186 of 186 batches, 1{,}111 of 1{,}113 programs; 64 $\to$ 0 \\
SWE-ReX's launch & host wrapper & command split and rejoined with \code{;} & a file sourced as one statement & 9 of 11 solutions \\
OpenHands's launch & host wrapper & special characters escaped & a file sourced as one statement & 34 of 34 commands \\
\bottomrule
\end{tabularx}

\smallskip
{\footnotesize\raggedright Effect: calls rerouted, lost pairs recovered, changed calls before and with \system ($n \to 0$).\par}
\end{table}

A rendering carries the call through a channel that the named hop does not parse.
A hop changes a call by parsing its text.
More quotes or escapes only move the change to the next parse, whereas an unparsed channel leaves the hop nothing to change.
The program then receives the intended action, which we call the \emph{rendering condition}.
The channels are a sourced file, an argument vector, standard input, and an encoded argument that the host wrapper passes on unparsed.
Where the channel is given, as PowerShell~5.1 builds the command line of a native program, the rendering writes each constant argument $s$ of a native invocation by the receiver's rules,
\begin{equation}
\label{eq:native}
\mathrm{native}(s) =
\begin{cases}
\text{\texttt{""}} & s = \varepsilon,\\
\text{\texttt{"}}\,\mathrm{esc}(s)\,\beta(s)\,\text{\texttt{"}} & s \text{ contains white space},\\
\mathrm{esc}(s) & \text{otherwise},
\end{cases}
\end{equation}
where $\mathrm{esc}$ writes \code{\textbackslash"} for a double quote and doubles the run of backslashes before it, and $\beta(s)$ repeats the run of backslashes that ends $s$, so that the closing quote stays a delimiter.
PowerShell~5.1 adds the quotes of the second case itself, and where it adds none, the rendering writes its own quotes and doubles each quote inside the value.
The C runtime of the program parses $\mathrm{native}(s)$ back to $s$.
For example, \code{print("hi")} becomes \code{print(\textbackslash"hi\textbackslash")} (Figure~\ref{fig:method}).
A launch that changes calls at 2 hops receives both renderings, that of the later hop first.
On the appended launch, the script is rendered for the process interface and then encoded for the wrapper, which consumes the quotes first.
\system refuses a call that no rendering carries, and names the hop.
Executing such a call would run a different action without notice, so the refusal is the only outcome that keeps the correspondence.
The \emph{domain} of a rendering is the set of texts that its channel carries, and \code{cmd.exe}, which ends a command at its first line break, carries no multi-line text.
The refused call returns a tool result that states what was not executed and why.

\begin{algorithm}[htbp]
\caption{\system: repair at the named hop.}
\label{alg:intact}
\footnotesize
\begin{algorithmic}[1]
\Require path $\Pi$ of the launch, its named hops $H$, text $x$
\Ensure a text $x'$ that satisfies the rendering condition, or a refusal that names the hop
\If{$x$ is outside the domain of the renderings for $H$} \Return $\mathrm{refuse}(H)$ \Comment{\code{cmd.exe} line break, Git Bash CRLF, target-parser fault}
\EndIf
\State $x' \gets x$
\If{the process interface $\in H$} \Comment{one-argument and appended launches; nothing is executed}
  \State $T \gets \mathrm{ParsePowerShell}(x')$; each constant argument $s$ of a native invocation becomes $\mathrm{native}(s)$ (Eq.~\ref{eq:native}); $x' \gets \mathrm{Unparse}(T)$
\EndIf
\If{the host wrapper $\in H$}
  \If{the launch appends the text after \code{-Command}} \State $x' \gets$ \code{-EncodedCommand }$\mathrm{Base64}(\mathrm{UTF16}(x'))$ \Comment{appended launch: rendered, then encoded}
  \Else \State $f \gets$ a file that holds $x'$ with its line ends; $x' \gets$ \code{source }$f$ \Comment{Claude Code, SWE-ReX, OpenHands}
  \EndIf
\EndIf
\If{the shell parse $\in H$ and the launch types into an interactive Bash}
  \State $x' \gets$ the typed line that switches off history expansion and completion, then $x'$
\EndIf
\State \Return $x'$
\end{algorithmic}
\end{algorithm}

A rendering changes the channel and keeps the launch.
The repair must not change the call itself, so the agent's action, the working directory, and the environment stay as the harness set them.
The sourced file runs in the shell process that would have evaluated the text, so the working directory, the exports, the exit status, and Claude Code's wrapper are those of the original launch.
\system selects the rendering by $\Pi$, $H$, and $x$ without a further measurement per call.
Section~\ref{sec:rq4} compares this selection with repairs that omit it.

\section{Experimental Setup}
\label{sec:study}

\subsection{Research Questions}
\begin{itemize}[noitemsep,topsep=0pt,leftmargin=*]
\item \textbf{RQ-1: How often, and at which hop, does a call change in real-world use?}
We replay the recorded shell calls of production sessions, public runs, and MCP servers through the witnesses of their hops on Windows and Linux.
\item \textbf{RQ-2: How much does the path change the score, recovery, and completion claims?}
We run \iecbench under 6 execution paths with 4 LLMs and, on tasks that differ in one literal, count the steps, tokens, and completion claims of an agent after a changed call.
\item \textbf{RQ-3: Does the IEC protocol attribute a failure to the path correctly?}
We compare the IEC protocol's attribution with published techniques and ablations against human labels and the repair replay.
\item \textbf{RQ-4: Does \system recover the loss?}
We inject faults at known hops and apply \system with the recorded actions held fixed and with the agent in the loop, against published interventions and ablations.
\end{itemize}

\subsection{Real-World Corpora}
The Windows corpus contains 47{,}828 shell calls that Claude Code and Codex ran through Git Bash and Windows PowerShell~5.1 in 261 sessions by 6 developers from a company over 11 weeks.
No session was created for this study.
A session is a conversation with its subagents and resumed continuations.
A resumed conversation copies its earlier calls into the new transcript, and we count each call once.
We exclude the project of this study.
We study the \emph{exposed calls} of the corpus, which we select by their text before execution.
A Bash call is exposed when it carries a heredoc, a backslash pair, or at least 1{,}000 characters.
A PowerShell call is exposed when it invokes a native program with constant arguments, one of which contains a double quote, a backslash, whitespace, or is empty, the only argument forms that the process interface changes.
The corpus has 11{,}982 exposed calls in 199 sessions, and in the replay no Bash call outside this set changes.
The Linux corpus contains the Codex and Claude Code shell calls of 9 developers on a shared host, in 71 sessions over 8 months.
The witness ran on the host at collection time with the developers' consent, and only its verdicts left the host, with hashed users and sessions and no command text.
2 external developers contributed held-out sessions with 1{,}921 calls on Windows and 161{,}214 on Linux, to which we apply the same witnesses and rules.

We also analyze public runs and datasets.
The runs include the Terminal-Bench~2 runs of Qwen Code, OpenCode, Claude Code, Codex, and OpenHands, and the Toolathlon~\citep{li2026toolathlon} trajectories for 22 LLMs on 108 tasks.
They also include keystroke logs from the Terminus-2 and Terminus-KIRA submissions to the Terminal-Bench~2 leaderboard, plus 203{,}710 public SWE-agent, mini-SWE-agent, and OpenHands trajectories.
4 public datasets provide the text a call would carry: the 89 reference solutions of Terminal-Bench~2\footnote{\url{https://github.com/harbor-framework/terminal-bench-2}}, the terminal commands of Toolathlon, the PowerShell commands of CLI-1M\footnote{\url{https://huggingface.co/datasets/carosh/cli-1m}}, and the argument values of ToolHop~\citep{ye2025toolhop}.

\subsection{\iecbench}
\iecbench has 67 pattern pairs, 10 multi-line programs, 135 ToolHop chains, and 46 tasks of Terminal-Bench~2.
Each hazard class of \iecbench is a change observed in the recorded sessions and runs, which we write with new literals and data so that no session text is released.
A \emph{pattern pair} is a hazard variant and a control variant of a task that require the same tool calls and differ in one literal and the input data built from it.
The literal triggers the change only in the hazard variant.
The pair is \emph{lost} when the control variant passes and the hazard variant fails.
An agent runs the pattern pairs with a shell tool alone, \emph{shell only}, or with a file-write tool in addition.
The 135 chains, of 3 to 7 steps, are up to 30 per step count drawn from the 990 ToolHop queries that pass a value requiring quoting, plus 5 that pass none.
We pair each Terminal-Bench task across a Linux and a Windows container with unchanged instruction, inputs, and verifier, and keep it only if the reference solution passes on both.
The oracles check the state that a task leaves against expected values computed without a shell.
\iecbench runs under 6 \emph{transports}, its execution paths.
2 are references: direct JSON arguments, and Git Bash with the command in a file, the control.
2 reproduce Claude Code's Bash tool and Qwen Code's \code{cmd.exe /d /s /c}.
The other 2 are the PowerShell~5.1 launches of Section~\ref{sec:witness}, the one-argument launch as in DeepSeek Harness and Codex, and the appended launch as in Aider.
4 LLMs run \iecbench: qwen3-coder-plus through its API, and Qwen2.5-Coder-32B, Qwen2.5-72B, and Hermes-3-Llama-3.1-70B (Hermes-3) on vLLM.
The 3 LLMs on vLLM also run the repair comparison of RQ-4.

2 further tasks, a fix task and a commit task, measure the recovery after a changed call.
15 LLMs from 4 vendors drive an agent with 2 tools, a structured file write and a command, on these tasks, which depend on one literal.
In the \emph{fix task}, the agent repairs a program and runs its test file, \code{test\_app (v2) \& more.py}.
In the \emph{commit task}, run by the 13 LLMs served through an API, the literal is a commit message that contains quotes.
Each LLM runs each task 3 times at temperature 0.6, with at most 8 steps in the fix task and 6 in the commit task.

\subsection{Baselines and Ablations}
\label{sec:baselines}
For attribution, we compare the IEC protocol with 3 LLM judges, 2 replay techniques, and 4 ablations.
The judges are those of Who\&When~\citep{zhang2025whowhen}: all-at-once, step-by-step, and binary search.
They run with their released prompts and settings on Qwen2.5-72B, an LLM from the original evaluation, without ground truth, up to the failed call.
We map a verdict that names the assistant to generation, and one that names a tool to the path.
The 2 replay techniques run on the lost pairs of \iecbench.
Causal Agent Replay~\citep{shah2026car} resamples the run at each agent step with 8 rollouts, and QuoteBench's fixed-reply replay~\citep{li2026quotebench} sends the recorded actions through a temporary file.
The QuoteBench reference is the repair replay on the 324 lost pairs that this replay attributes, the 137 recovered pairs, and the 187 without a changed call in Section~\ref{sec:rq4}.
The reference of Causal Agent Replay is the witness on the shell-only losses.
4 ablations each remove one component of the IEC protocol: no attribution, no receiver parse, no witness, and an LLM judge in place of the witness and the receiver parse, run with 2 LLMs.

We score the judges and the ablations against a human reference from 2 samples of the production sessions, drawn at random within each verdict of the IEC protocol.
The first holds 117 loud failures labeled generation, path, or nested, and the second 120 pairs of a sent and a received text labeled by whether they have the same effect.
2 annotators labeled both samples following written guidelines and without this verdict, with $\kappa$ of 0.79 and 0.89.
The reference is the label on which both agree, for 102 failures and 115 pairs.
Execution evidence replaces this label in 13 items.

For repair, we compare \system on the same items with 4 published interventions and 3 ablations.
2 come from QuoteBench~\citep{li2026quotebench}: the temporary file and the disclosed-boundary contract, a description of the transport's path in place of the sentence that names the shell.
The third is the structured execution feedback of Gupta et al.~\citep{gupta2026dissecting}, with command, status, exit code, output, and error in the tool result.
The fourth is Reflexion~\citep{shinn2023reflexion}, which reflects on a failed trial and retries, using the task oracle as the evaluator.
The ablations are each of \system's 2 repairs alone on the appended launch, and an evidence ablation that reports the call as sent and as received, with the named hop, in the tool result.

\subsection{Metrics}
\label{sec:metrics}
\noindent\textbf{Changed calls, silent calls, changed-call rate} (Tables~\ref{tab:matrix} to~\ref{tab:mcp}, Figure~\ref{fig:wild}).
A call $c$ denotes an action $a(c)$ and the receiver executes $r(c)$ (Section~\ref{sec:model}).
The call is a \emph{changed call} when $r(c) \neq a(c)$, and silent as defined in Section~\ref{sec:model}.
The \emph{changed-call rate} of a set of calls is the share of its calls that are changed.

\noindent\textbf{Arrives as written, rerouted, refused, time added per call} (Tables~\ref{tab:intact} and~\ref{tab:forms}).
A call \emph{arrives as written} when it is not changed.
A deployment \emph{reroutes} a call when it delivers the call through a rendering, and \emph{refuses} a call that no rendering carries.
The time added per call is the median wall-clock time that the form adds.

\noindent\textbf{Pass@k} (Table~\ref{tab:iec-repair}).
An item is a hazard or control variant of a pattern pair, judged by its oracle, or a chain, judged by its answer.
A cell reports pass@$k$, the number of its items that pass within $k$ trials.
Every cell is pass@1, 1 trial at temperature 0, except the 2 columns with 10 trials of Table~\ref{tab:iec-repair}.
A pair allows 10 steps and 120~s per command, and a chain 15 steps and 60~s.
The 2 columns run Reflexion with 10 trials, the task oracle as evaluator, and the same limits on the original launch and on the \system transport, so the path is the only difference.

\noindent\textbf{Tokens per passed item, cost ratio} (Section~\ref{sec:rq2}).
The \emph{tokens per passed item} are the tokens of all items in a cell divided by the items that pass.
The \emph{cost ratio} of a launch and a task is this value on the original launch over its value with \system.

\noindent\textbf{Recovered pairs} (Table~\ref{tab:intact}).
A lost pair is \emph{recovered}, a \emph{path-caused loss}, when its recorded actions pass once the named hop is repaired, with every action held fixed.

\noindent\textbf{Effect of a repair} (Tables~\ref{tab:intact} and~\ref{tab:iec-repair}).
The effect compares pass@$k$ with and without the repair on the same items, and the calls that the launch changes with and without the repair.
The $b$ items that pass only with the repair and the $c$ items that pass only without give the exact McNemar test, per LLM and pooled over the 2 Qwen2.5 LLMs.

\noindent\textbf{Agreement, $\kappa$, attributed to the path, tokens} (Table~\ref{tab:attr-baselines}).
Per failure, a technique and the reference each assign generation, path, or nested.
\emph{Agreement} is the share of failures with the same label, and Cohen's $\kappa$ corrects it for chance.
\emph{Attributed to the path correctly} counts the failures that both attribute to the path, and \emph{incorrectly} those that only the technique does.
\emph{Tokens} is the number of LLM tokens per failure.
A replay technique also reports the share of the path-caused losses it detects.

\noindent\textbf{Localization, steps, tokens, retries, false confirmation} (Sections~\ref{sec:rq4} and~\ref{sec:rq2}).
Per injected fault, we report whether the first divergence names the hop of the fault and whether \system restores or refuses the call.
Per episode with the agent in the loop, we report the outcome, steps, tokens, retries, and a \emph{false confirmation}, a completion claim on a failed run.

\section{Experiments \& Results}
\label{sec:evaluation}

\subsection{RQ-1: How Often, and at Which Hop, Does a Call Change in Real-World Use?}
\label{sec:rq1}
\noindent\textbf{[Objective and Design]:} To measure how often, and at which hop, a call is executed as a different action, we replay each recorded call through the witness of its hop.
The replay uses the shortest wrapper that the harness generates, and we compare the received action with the sent one (Sections~\ref{sec:witness} and~\ref{sec:fd}).
In the last 3 weeks of the Windows corpus, \system delivers the Bash commands as files.
On Linux, the corpora and 99 shell blocks of the anthropics/skills repository\footnote{\url{https://github.com/anthropics/skills}} pass through the released command handling of OpenHands, Gemini CLI, Aider, and SWE-ReX, Terminus-2's tmux typing, and dash and zsh.
The harnesses and 18 MCP servers that execute commands receive 20 payloads under an argv sentry.
The payloads are argument values with quotes, backslashes, line breaks, and other special characters.
Each harness runs in its default configuration on Windows~11 with Windows PowerShell~5.1 and Git for Windows.
4 structured MCP servers receive the payloads as JSON arguments, and we compare what the backend stores with the contract of each tool.

\begin{figure}[htbp]
\centering
\begin{subfigure}[t]{0.33\textwidth}\centering
\includegraphics{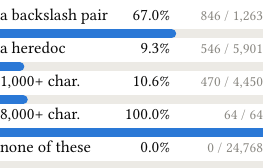}
\caption{Changed calls, by feature.}\label{fig:wild-a}
\end{subfigure}\hfill
\begin{subfigure}[t]{0.33\textwidth}\centering
\includegraphics{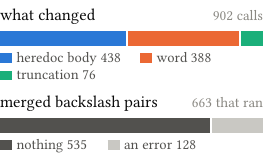}
\caption{Kind of change, feedback.}\label{fig:wild-c}
\end{subfigure}\hfill
\begin{subfigure}[t]{0.32\textwidth}\centering
\includegraphics{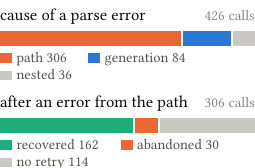}
\caption{Parse errors, cause and recovery.}\label{fig:wild-d}
\end{subfigure}
\caption{Changed calls and parse errors in the production sessions.}
\label{fig:wild}
\end{figure}

\noindent\textbf{[Experimental Results]:}
In the replay, 1{,}199 of the 11{,}982 exposed calls change, a changed-call rate of 10.0\% (Table~\ref{tab:matrix}).
Through Claude Code's Bash launch, 12.0\% of the 7{,}491 exposed Bash calls change.
The rate is 67.0\% for a call with a backslash pair and 100\% for a call of at least 8{,}000 characters (Figure~\ref{fig:wild-a}).
The wrapper changes a Bash call in 2 ways.
It merges backslash pairs, and the wrong action runs without any reported error for 535 of the 663 such calls that ran before \system, so 80.7\% of these changes are silent (Figure~\ref{fig:wild-c}).
It also cuts the command, and all 11 cut calls that ran before \system fail with an error.
\system rerouted 227 of the changed calls, and 1 was rejected at Claude Code's approval prompt before it ran.
A changed call occurs in 65 of the 199 sessions with an exposed call, a share of 32.7\%.
On the Bash tool, 30 of the 39 sessions with an exposed call contain one, and 23 of the 27 sessions with at least 100 Bash calls.

The 10 harnesses change calls at 3 of the 5 hops, and the hop differs between harnesses and between platforms (Table~\ref{tab:matrix}).
The PowerShell paths deliver the script intact and change the call at the process interface, where PowerShell~5.1 passes arguments to a native program.
196 of the 297 changed calls lose a double quote within a value, and 127 receive a different number of arguments.
The same hop changes payloads in \code{wsl.exe}, the batch shims of npm, npx, and yarn, Gemini CLI's Windows launch, and an interactive PowerShell.
Every payload that a harness changes is reported as a success.
Under Qwen Code's \code{cmd.exe} launch, 30 of the 31 multi-line commands run as an empty program.
On Linux, 4 harnesses change none in 80 runs.
The held-out sessions of the 2 external developers show the same 2 kinds of change at the same hop, in 5 of 605 and 2 of 52 exposed calls, all silent.

\begin{table}[htbp]
\centering
\caption{The hop at which each harness changes a call.}
\label{tab:matrix}
\footnotesize
\setlength{\tabcolsep}{4pt}
\begin{tabular}{@{}lcccccc@{}}
\toprule
 & \multicolumn{3}{c}{Windows} & \multicolumn{3}{c}{Linux} \\
\cmidrule(lr){2-4}\cmidrule(l){5-7}
Harness & \shortstack{\ding{183}~host\\wrapper} & \shortstack{\ding{184}~shell\\parse} & \shortstack{\ding{185}~process\\interface} & \shortstack{\ding{183}~host\\wrapper} & \shortstack{\ding{184}~shell\\parse} & \shortstack{\ding{185}~process\\interface} \\
\midrule
Claude Code & \ding{108}~902/7{,}491 & & \ding{108}~125/2{,}407 & \multicolumn{3}{c}{none} \\
Codex & & & \ding{108}~172/2{,}084 & \multicolumn{3}{c}{none} \\
Qwen Code & & \ding{108}~30/31 & & \multicolumn{3}{c}{none} \\
OpenCode & & & \ding{108}~12/99 & \multicolumn{3}{c}{none} \\
DeepSeek Harness & & & \ding{108}~2/20 & \multicolumn{3}{c}{none} \\
Gemini CLI & & & \ding{108}~2/20 & \multicolumn{3}{c}{none} \\
Aider & & \ding{108}~4/20 & & & \ding{108}~72 & \\
OpenHands & \multicolumn{3}{c}{n/a} & \ding{108}~34 & & \\
SWE-ReX & \multicolumn{3}{c}{n/a} & \ding{108}~13 & & \\
Terminus-2 & \multicolumn{3}{c}{n/a} & & \ding{108}~186/245 & \\
\bottomrule
\end{tabular}

\smallskip
{\footnotesize\raggedright \ding{108}: changed over exposed calls, payloads, or batches. Hops numbered as in Figure~\ref{fig:method}.\par}
\end{table}

\begin{table}[htbp]
\centering
\footnotesize
\setlength{\tabcolsep}{5pt}
\caption{A command on Linux before Bash parses it.}
\label{tab:linux}
\begin{tabular}{@{}lrr@{}}
\toprule
 & \multicolumn{2}{c}{Commands changed} \\
\cmidrule(l){2-3}
Handling of the command & silently & with an error \\
\midrule
An argument to \code{bash -c} or \code{bash -lc} & 0 & 0 \\
Terminus-2 types it into an interactive Bash & 4 & 3 \\
SWE-ReX splits it and rejoins it with \code{;} & 12 & 1 \\
OpenHands escapes it & 29 & 5 \\
dash parses it, where \code{/bin/sh} is dash & 45 & 27 \\
\bottomrule
\end{tabular}
\end{table}

On Linux, a call changes only where the harness handles the command before Bash parses it.
A command passed to its shell as an argument arrives as written in all 1{,}779 calls of 4 harnesses, 4{,}012 calls of 9 developers, and 161{,}214 held-out Codex calls.
We measure the other handlings on 5 corpora without an LLM (Table~\ref{tab:linux}).
Terminus-2 types the command into an interactive Bash, where completion silently replaces a tab and history expansion drops a line with a double-quoted \code{!x}.
1{,}113 of the 1{,}118 Toolathlon programs with an expandable \code{!} change when typed as \code{python3 -c}, against 2 of the 12{,}030 without a tab or such a \code{!}.
Aider, whose prompt asks the LLM for Bash, and 2 MCP servers pass the command to \code{/bin/sh}.
There, dash stops 5 reference solutions, which Terminal-Bench runs with Bash, at \code{set -o pipefail}, and zsh, the default shell of macOS, silently changes 5.

The changes to typed commands also occur in public runs.
In 4{,}434 leaderboard trials of Terminus-2 and Terminus-KIRA, 245 keystroke batches contain a tab or an expandable \code{!}.
186 of them change when typed through the witness, 81 at a tab, and 105 by history expansion, in 137 of the 4{,}356 scored trials.
In 203{,}710 public trajectories, only the harnesses that type into an interactive Bash produce \code{event not found}.
The error occurs 2{,}546 times in 1{,}532 of their 113{,}674 trajectories, and never in the 90{,}036 trajectories of the harnesses that pass the command to a non-interactive Bash.

A changed call that fails loudly misleads the agent about the cause.
Of the 47{,}828 shell calls, 709 contain a parser or quoting error message, 426 of them a Bash or Python parse error.
Of these 426, the IEC protocol attributes 306 to the path, 84 to generation, and 36 to a nested hop (Figure~\ref{fig:wild-d}).
The 306 are 253 cut commands, 46 merged backslash pairs, and 7 quotes left open at the wrapper.
Each of the 306 was reported to the agent as an error in its own text.
In 231 of the 306, the shortest wrapper delivers the call as written and only the longer wrapper of its session changed it, so the replay's changed-call rate is a lower bound.

\begin{table}[htbp]
\centering
\footnotesize
\setlength{\tabcolsep}{6pt}
\caption{The 16 MCP servers that receive 20 payloads.}
\label{tab:mcp}
\begin{tabular}{@{}lrrrrr@{}}
\toprule
 & & \multicolumn{2}{c}{Windows} & \multicolumn{2}{c}{Linux} \\
\cmidrule(lr){3-4}\cmidrule(l){5-6}
The server passes the call to & Servers & run & change a payload & run & change a payload \\
\midrule
a shell, as one string & 6 & 6 & 4 & 4 & 0 \\
a program, as an argument vector or a file & 3 & 2 & 0 & 2 & 1 \\
\code{git}, as a commit message & 3 & 3 & 0 & 3 & 0 \\
its own backend, as JSON values & 4 & 4 & 1 & 4 & 1 \\
\bottomrule
\end{tabular}
\end{table}

An agent that reaches a shell through an MCP server inherits the launch that the server constructs.
Table~\ref{tab:mcp} covers 16 of the 22 servers, 12 of the 18 that execute commands and the 4 structured servers.
Of the other 6, 5 type into a terminal and 1 passes an argument vector.
6 of the 16 and 4 of the 6 change a payload.
A server also changes the text itself.
The fetch server turns a backslash in a URL into a slash, and the SQLite server describes a nonexistent table when the name needs quoting.
The file-system server writes \code{a:b} into an alternate data stream of \code{a} on Windows.
Each change is reported as a success, and the tool description mentions none.

\find{{\bf [RQ-1] Findings:} (1)~In production, 10.0\% of the exposed calls change, 12.0\% at the host wrapper of Claude Code's Bash launch, and 6.6\% at the process interface of the PowerShell launches. For 80.7\% of the Bash calls whose backslash pairs the wrapper merges, the wrong action runs without any reported error, and a cut call fails with an error.
(2)~Of the 426 calls that failed with a parse error, 306 carried valid text that the path changed, and all 10 harnesses and 10 of the 22 MCP servers changed a call.
{\bf Insight:} A changed call comes from the launch, not from the platform or the LLM, so a harness should be tested hop-by-hop.}

\subsection{RQ-2: How Much Does the Path Change the Score, Recovery, and Completion Claims?}
\label{sec:rq2}
\noindent\textbf{[Objective and Design]:} To quantify the path's effect on score, recovery after a changed call, and the agent's final report, we run \iecbench under 6 transports with 4 LLMs.
We hold each task fixed while one literal varies, and count the steps, the tokens, and the completion claims on failed runs.
In the fix and commit tasks, the launch appends the agent's command after \code{powershell.exe -Command}.
6 harnesses run 10 Terminal-Bench~2 pairs through Harbor\footnote{\url{https://doi.org/10.5281/zenodo.20953922}} with qwen3-coder-plus, and Qwen Code\footnote{\url{https://github.com/QwenLM/qwen-code}} on Windows receives the instruction as an argument or as a file.
All 46 pairs' instructions pass through both deliveries without an LLM.
For the 306 path-caused failures of the production sessions, we count the steps and tokens until the agent recovers or abandons the call.

\begin{figure}[htbp]
\centering
\includegraphics{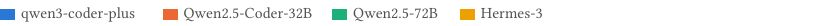}

\includegraphics{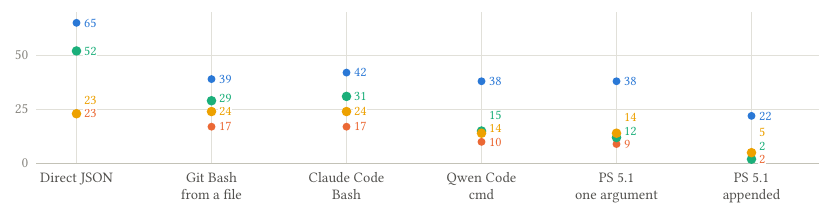}
\caption{Chains answered correctly, of 135, under 6 transports.}
\label{fig:chains}
\end{figure}

\noindent\textbf{[Experimental Results]:}
On the 135 chains, all but 4 changed calls occur on the 2 PowerShell launches.
The 4 LLMs, in the order of Figure~\ref{fig:chains}, encounter a changed call in 16, 71, 63, and 34 of the 135 chains on the one-argument launch and in 128 to 133 on the appended launch.
On the appended launch, 737 of the 1{,}165 calls of qwen3-coder-plus change, 662 at a value with a space.

A score combines 3 effects (Figure~\ref{fig:chains} and Table~\ref{tab:iec-repair}).
First, on the control transport, which changes no calls, the 3 Qwen LLMs answer 6 to 26 fewer chains than with direct JSON calls.
Hermes-3 loses none.
Second, the PowerShell launches reduce the score further, to as few as 2 chains.
Hermes-3, 29 chains behind Qwen2.5-72B with direct JSON calls, is then 2 ahead on the one-argument launch.
Third, repairing the hop with \system raises the 2 Qwen2.5 LLMs by 12 and 17 chains on the one-argument launch, to or above the control level.
Qwen2.5-72B is then 5 and 4 chains ahead of Hermes-3 again, as on the control.

A changed call usually breaks a chain at its first step.
On the appended launch, the first changed call is the first tool call in 483 of the 524 affected chain runs of the 4 LLMs, and the tool's argument parser rejects it in 492.
After a changed call, the LLMs send the command again or rewrite it, leaving out the part of the value that the hop split off.

After a changed call, an LLM revises a command that was correct.
In the fix task, all 13 LLMs served through an API pass \code{"test\_app (v2) \& more.py"} to pytest in their first command.
The command is valid PowerShell, and the appended launch converts it into an \code{AmpersandNotAllowed} error.
The 2 local LLMs write the name unquoted.
In all 45 runs, the first command fails, and the run takes 2 to 7 steps in total.
Under PowerShell~5.1, an LLM uses a median of 1.8 times as many tokens as in its Linux run.
In the commit task, all 13 LLMs write a correct \code{git commit} command.
The commit carries the intended message in 1 of 39 runs, and 19 runs end in a false confirmation.

In the production sessions, the agent recovered 162 of the 306 path-caused failures by retrying, in 2 to 8 steps.
It abandoned 30 after retries and did not retry 114 (Figure~\ref{fig:wild-d}).
Each call that finally succeeded cost 3.9 times the tokens of one clean request.

Over all rows of Table~\ref{tab:iec-repair}, a passed task costs 2.4 times as many tokens on the original launch as with the call delivered by \system.
The cost ratio is 12.3 for the chains and 7.9 for the pairs on the appended launch, and 4.0 on Claude Code's Bash launch.

On the pattern pairs, whose variants differ in one literal, the hazard variant passes far less often on a launch that changes calls.
Without the write tool, it passes in 4 to 11 of the 67 pairs on the appended launch, against 38 to 46 on the control transport.
In this setting, every empty-argument hazard fails on both PowerShell launches, every truncation hazard fails on Claude Code's Bash launch, and none of the 10 multi-line programs pass on the appended launch.
A changed call occurs in 70 lost pairs on the one-argument launch and in 81 on the appended launch.
On Claude Code's Bash launch, 13 of the 22 lost pairs with a changed call involve a truncated command.
Agents report 665 of the 949 failed runs of hazard variants and multi-line programs as completed, a false-confirmation rate of 70.1\%.

The delivery of the instruction to the agent also determines a score.
Delivered as a command-line argument, 33 of the 46 instructions are cut at their first line break, 3 lose all their double quotes, and 5 exceed the command-line limit of \code{cmd.exe}.
Delivered as a file, all 46 arrive as written.
On the 10 pairs with agent runs, every run with the argument scores 0, and with the file Qwen Code passes 6 of 30 runs on Windows against 7 of 30 on Linux.

\find{{\bf [RQ-2] Findings:} (1)~The path removes an LLM's lead: Qwen2.5-72B, 29 chains ahead of Hermes-3 with direct JSON calls, is 2 and 3 chains behind on the PowerShell launches, and 5 and 4 ahead again once \system repairs them.
(2)~After a changed call, the agent rewrites its own correct call. A passed task costs 2.4 times the tokens, and up to 12.3 times for the chains of the appended launch. 1 of 39 runs commits the intended message. In production, 144 of the 306 path-caused failures are not recovered by a retry. Agents report 665 of 949 failed runs as completed.
{\bf Insight:} A score should reflect the LLM, not the launch, so a benchmark should report its launch and hold it fixed.}

\subsection{RQ-3: Does the IEC Protocol Attribute a Failure to the Path Correctly?}
\label{sec:rq3}
\noindent\textbf{[Objective and Design]:} To evaluate the IEC protocol's attribution, we compare it with the Who\&When judges, 2 replay techniques, and 4 ablations (Section~\ref{sec:baselines}).
The references include human labels for 102 failures and 115 pairs, plus the repair replay.

\begin{table}[htbp]
\centering
\caption{Attribution of 102 failures.}
\label{tab:attr-baselines}
\footnotesize
\setlength{\tabcolsep}{5pt}
\begin{tabular}{@{}lrrrr@{}}
\toprule
 & \multicolumn{2}{c}{Attributed to the path} & & LLM tokens \\
\cmidrule(lr){2-3}
Approach & correctly (of 56) & incorrectly (of 46) & $\kappa$ & per failure \\
\midrule
\multicolumn{5}{c}{\emph{Baselines: published techniques}} \\
Who\&When all-at-once~\citep{zhang2025whowhen} & 0 & 0 & 0.00 & 20.4k \\
Who\&When step-by-step~\citep{zhang2025whowhen} & 29 & 20 & 0.04 & 37.8k \\
Who\&When binary search~\citep{zhang2025whowhen} & 25 & 20 & 0.01 & 42.6k \\
\multicolumn{5}{c}{\emph{Baselines: ablations of the IEC protocol}} \\
LLM judge, qwen3-coder-plus & 8 & 0 & 0.10 & 0.7k \\
LLM judge, Qwen2.5-72B & 2 & 1 & 0.02 & 0.6k \\
No attribution (end-to-end score) & 0 & 0 & 0.00 & 0 \\
No receiver parse (string comparison) & 13 & 5 & 0.09 & 0 \\
No witness or parse check (error message alone) & 56 & 46 & 0.00 & 0 \\
\midrule
\textbf{IEC protocol (ours)} & \textbf{54} & \textbf{1} & \textbf{0.77} & \textbf{0} \\
\bottomrule
\end{tabular}
\end{table}

\begin{table}[htbp]
\centering
\caption{Path-caused losses that one reference path detects, by hop.}
\label{tab:refpath}
\footnotesize
\setlength{\tabcolsep}{6pt}
\begin{tabular}{@{}llrr@{}}
\toprule
Hop that changed the call & Launch & Path-caused losses & Detected~\citep{li2026quotebench} \\
\midrule
host wrapper & Claude Code's Bash launch & 20 & 20 \\
process interface & One-argument launch & 48 & 0 \\
host wrapper, then process interface & Appended launch & 69 & 31 \\
\midrule
All & & 137 & 51 (37.2\%) \\
\bottomrule
\end{tabular}
\end{table}

As Table~\ref{tab:attr-baselines} shows, the IEC protocol agrees with the reference on failure attribution at $\kappa$ 0.77, whereas the all-at-once judge attributes 97 of the 102 failures to the LLM.
The IEC protocol calls no LLM, and a Bash witness takes 0.1~s per call.
On the action change, the IEC protocol agrees with the reference on all 115 pairs, with $\kappa$ 1.00.
String comparison and the qwen3-coder-plus judge, which read the same sent and received texts, reach $\kappa$ 0.65 and classify 15 of the 35 unchanged pairs as changed.

The replay techniques detect a path-caused loss only where their intervention reaches the hop that changed the call.
Causal Agent Replay asks at which step the outcome was determined, and a change within a call is not a step.
Resampling a step gives a passing rollout in 12 of the 67 shell-only lost pairs with a changed call and in 29 of the 53 without, so a rescue does not indicate the path.
On the 324 lost pairs of \iecbench that the repair replay attributes, QuoteBench's fixed-reply replay agrees with the repair replay at 0.73, with $\kappa$ 0.41.
The replay of QuoteBench attributes no lost pair incorrectly to the path and detects 37.2\% of the path-caused losses (Table~\ref{tab:refpath}).
It detects every loss at the wrapper and none at the process interface, and on the appended launch only the 31 that the wrapper repair alone recovers.
Its reference path, a temporary file, bypasses the wrapper and goes through the same native argument passing.

\find{{\bf [RQ-3] Findings:} (1)~The IEC protocol agrees with the human labels of 102 failures, 56 of them caused by the path, at $\kappa$ 0.77, whereas the all-at-once judge attributes 97 to the LLM, and the Who\&When judges reach $\kappa$ at most 0.04.
(2)~A reference path detects only the hop it bypasses: all 20 losses at the wrapper, none of the 48 at the process interface.
{\bf Insight:} The trajectory does not show what a hop received, so failure analysis should check the path before it attributes a failure to the LLM.}

\subsection{RQ-4: Does \texorpdfstring{\system}{\systemname} Recover the Loss?}
\label{sec:rq4}
\noindent\textbf{[Objective and Design]:}\label{app:dataset} The first divergence should name a fault injected at a known hop, and \system, which repairs only the named hop, should recover the losses that the path caused.
First, we test the localization by fault injection on Windows.
We start from a base path that delivers all 24 payloads unchanged, and inject one of 7 known faults at a time, each at one of the 5 hops.
The 24 payloads are the 20 base payloads and 4 further cases, a long string, a variable expression, trailing spaces, and quotes with a space, which gives 168 injections.
Second, we render the modified Linux batches and commands (Table~\ref{tab:linux}) with \system and pass them through the harness's release handling and the Bash witness.
We also type 30 further pattern pairs, the typed pairs, into an interactive Bash as Terminus-2 does.
Their literal contains a \code{!} inside double quotes.
Third, on \iecbench, we replay the recorded writes and commands of a run in a fresh sandbox through the transport with the named hop repaired.
Where a launch changes calls at 2 hops, we repair both, and the oracle grades the final state.
This replay covers the 360 lost pairs whose failure the original transport reproduces: 173 with a changed call on the 3 launches that have a rendering, and 187 without.
Fourth, we evaluate the repairs with the agent in the loop.

\begin{figure}[htbp]
\centering
\begin{subfigure}[t]{0.46\textwidth}\centering
\includegraphics{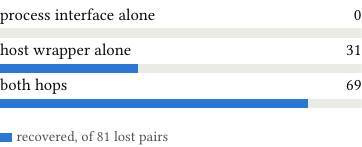}
\caption{Appended launch, by repaired hop.}\label{fig:repair-a}
\end{subfigure}\hfill
\begin{subfigure}[t]{0.46\textwidth}\centering
\includegraphics{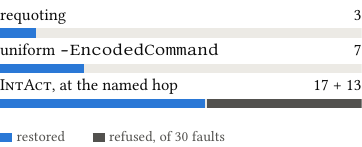}
\caption{Injected faults, by repair.}\label{fig:repair-b}
\end{subfigure}
\caption{A repair at the named hops against a repair at one hop.}
\label{fig:repair}
\end{figure}

\noindent\textbf{[Experimental Results]:}
Of 168 injections, 30 change what the target receives, and the first divergence names the hop in all 30 and in none of the other 138.
The 30 are 3 at the tool serialization, 8 at the wrapper, 3 at the shell parse, 3 at the process interface, and 13 at the target parser.
\system restores every call of the first 4 hops and refuses the 13 at the target parser, whereas requoting every argument or encoding every command restores a fraction of them (Figure~\ref{fig:repair-b}).
To test the comparison itself, we mutate command texts under a hop's text operators.
Of 297 mutants, Bash executes 262 differently from the original, and the comparison detects all 262.
On the Linux launches (Table~\ref{tab:intact}), \system delivers every changed leaderboard batch, all but 2 of the 1{,}113 changed Toolathlon programs, every OpenHands command, and 9 of the 11 SWE-ReX solutions as written.

\begin{table}[htbp]
\centering
\caption{Items passed by the 3 LLMs, by launch and intervention.}
\label{tab:iec-repair}
\footnotesize
\setlength{\tabcolsep}{3pt}
\begin{tabular}{@{}lrrrrrrr@{}}
\toprule
 & \multicolumn{5}{c}{1 trial} & \multicolumn{2}{c}{10 trials} \\
\cmidrule(lr){2-6}\cmidrule(l){7-8}
Launch, task (items) & None & \system & Temp.\ file~\citep{li2026quotebench} & Boundary~\citep{li2026quotebench} & Feedback~\citep{gupta2026dissecting} & Reflexion~\citep{shinn2023reflexion} & \system \\
\midrule
One-argument, pairs (201) & 104 & \textbf{129} & 103 & 85 & 101 & 148 & \textbf{172} \\
One-argument, chains (405) & 35 & \textbf{74} & 39 & 35 & 45 & 81 & \textbf{155} \\
Appended, pairs (201) & 26 & \textbf{127} & 103 & 25 & 23 & 50 & \textbf{181} \\
Appended, chains (405) & 9 & \textbf{74} & 39 & 5 & 8 & 36 & \textbf{150} \\
Claude Code Bash, pairs (201) & 109 & \textbf{125} & 125 & 112 & 104 & 131 & \textbf{162} \\
Typed Bash, Linux, pairs (90) & 74 & \textbf{86} & n/a & 79 & 79 & 86 & \textbf{88} \\
\bottomrule
\end{tabular}
\end{table}

With the recorded actions held fixed, 137 of the 173 lost pairs with a changed call pass once \system repairs the named hop, so the repair recovers 79.2\% of them.
The share ranges from 68.6\% on the one-argument launch to 90.9\% on Claude Code's Bash launch (Table~\ref{tab:intact}).
None of the 187 lost pairs without a changed call pass, so the repair acts on the path alone, with Fisher's exact test at $p<.001$.
On the appended launch, the process interface alone recovers nothing, and the wrapper alone less than half of what both hops recover (Figure~\ref{fig:repair-a}).
Of the 36 unrecovered pairs, 22 fail at an action that no hop changed, 12 were abandoned after the misleading feedback, and 2 use a construct outside the domain of the rendering.

With the agent in the loop, \system raises pass@1 for every LLM on every launch (Table~\ref{tab:iec-repair}).
On Claude Code's Bash launch, the temporary file is \system's rendering.
Pooled over the 2 Qwen2.5 LLMs, \system passes 20 more pairs and loses 1 on the one-argument launch, and 81 more and loses 5 on the appended launch.
It answers 30 more chains against 1 and 46 against 1, each at $p<.001$ by the exact McNemar test, and gains 13 pairs against 2 on Claude Code's Bash launch, at $p=.007$.
Hermes-3's gains on both PowerShell launches are significant in all 4 tests, with $p \leq .031$.
Either repair of the appended launch alone recovers less than both.
The native rendering alone recovers nothing, and \code{-EncodedCommand} alone 38.9\% to 87.5\% of what both gain.
The chains of Claude Code's Bash and \code{cmd.exe} launches contain no changed calls and have no row in the table, and a multi-line command under \code{cmd.exe} can only be refused.

Evidence about the change does not help the agent, and retries recover part of the loss at many times the cost.
The evidence ablation of \system returns the call as sent and as received in the tool result.
It leaves the Qwen2.5 chains at 7 and 12 on the one-argument launch and 0 and 3 on the appended launch.
None of these differs significantly from the original launch, with $p \geq .500$ by the exact McNemar test.
With the boundary disclosed, Qwen2.5-Coder-32B chains its commands with \code{\&\&}, which PowerShell~5.1 rejects, in each of the 15 pairs with a regular expression, and it loses each one.
It then reports completion without having run its program.
Reflexion gains 11.4\% to 55.0\% of what \system gains with 1 trial on the appended launch, at 463 to 1{,}180 retries per cell.
With the same 10 trials on the \system transport, the 3 LLMs take 22 to 913 retries per cell on Windows launches.
They pass more pairs and answer more chains than Reflexion on every Windows launch (Table~\ref{tab:iec-repair}).

On Linux, \system switches expansion and completion off before typing.
No call of the 30 typed pairs then changes, and all 12 lost pairs with a changed call pass, 7, 2, and 3 for the 3 LLMs.
Without \system, the Qwen2.5 LLMs encounter history expansion in 14 and 12 hazard episodes, reported as \code{event not found}, and Hermes-3 in 6.
The Qwen2.5 LLMs recover in 7 of the 14 and 10 of the 12 affected episodes, in 2 to 7 steps.
They take a median of 4 and 5 steps per episode, compared with 2 with \system, and 2.6 times as many tokens.
4 and 1 of these episodes end in a false confirmation, compared with none with \system.
With 10 trials, Qwen2.5-Coder-32B needs no retries with \system, whereas Reflexion reaches 29 pairs with 20 retries.

\find{{\bf [RQ-4] Findings:} (1)~With the recorded actions held fixed, \system recovers 137 of the 173 lost pairs with a changed call and, as a control, none of the 187 without one. On the appended launch, the process interface alone recovers 0 of 69 pairs, the wrapper alone 31, and both 69.
(2)~The agent cannot repair the path from inside. With the boundary disclosed, one LLM falls from 50 pairs to 36. At the same 10 trials, \system passes significantly more items than Reflexion.
{\bf Insight:} The harness, not the LLM, establishes intent-execution correspondence, so it should repair each named hop in path order and refuse what it cannot deliver.}

\section{Discussion: \texorpdfstring{\system}{\systemname} in Deployment}
\label{sec:rq5}
We deploy and release \system in 4 forms (Table~\ref{tab:forms}).
The Claude Code mod\footnote{\url{https://code.claude.com/docs/en/plugins/mods/overview}} ran on the production sessions' calls inside Claude Code, without starting a process.
The mod replaces a command longer than 200 characters, or one with a heredoc or a backslash pair, with a statement that sources a file, and keeps the first line of the command as a comment in the transcript.
The Windows execution kernel of a company's desktop agent renders each native argument before launch, puts the script in a file, and refuses multi-line text under \code{cmd.exe}.
A shell shim first on \code{PATH} renders the script for Codex, DeepSeek Harness, Gemini CLI, and OpenCode and starts Windows PowerShell~5.1.
The MCP server's single tool delivers the command by rendering the platform's launch.
Each form changes the channel and not the action, and a refusal names what was not executed.
The time added per call is a median over the 3{,}577 rerouted production calls for the mod and over 20 payloads for the shim and the server.

\begin{table}[H]
\centering
\caption{Forms of \system.}
\label{tab:forms}
\footnotesize
\setlength{\tabcolsep}{4pt}
\begin{tabularx}{\columnwidth}{@{}>{\raggedright\arraybackslash}p{0.19\columnwidth}>{\raggedright\arraybackslash}p{0.2\columnwidth}Y>{\raggedright\arraybackslash}p{0.12\columnwidth}@{}}
\toprule
Form & Harnesses & Effect & Time added per call \\
\midrule
Mod on the Bash tool & Claude Code & on the 3{,}577 rerouted production calls, all 227 that would have changed rerouted and none of the 3{,}540 that ran changed & 4~ms \\
Kernel of a company's desktop agent (Windows) & the agent, released to its users & 197 of 9{,}032 replayed native invocations changed before, none after & n/a \\
Shell shim first on \code{PATH} & Codex, DeepSeek Harness, Gemini CLI, OpenCode & 20 of 20 payloads arrive as written through each, 18 without the shim & 396~ms \\
MCP server & any MCP client & 20 of 20 payloads through PowerShell~5.1 and Linux, 19 through Git Bash & 5.9, 2.1, 32~ms \\
\bottomrule
\end{tabularx}
\end{table}

In the production sessions, \system rerouted all 227 calls that would have changed, and its rendering, computed for all 47{,}119 recorded calls of the 3 launches, refuses none.
The 3 released forms add 2.1 to 396~ms per call.

\section{Threats to Validity}
\label{sec:threats}

\paragraph{\textbf{Internal validity.}}
Agent runs vary between executions, so on \iecbench we fix the temperature at 0 and compare conditions on the same items with an exact McNemar test.
Transports differ in language and in hops, so we measure a hop by its repair within the same launch.
A retrying baseline could gain from its extra trials alone, so the 2 columns with 10 trials run the same 10 trials, oracle, and limits on the original launch and on the \system transport.

\paragraph{\textbf{Construct validity.}}
A witness could change the call itself, so it adds nothing to the argument.
It uses an environment variable for Bash, a parser without execution for PowerShell, an argv probe, or a stub in place of the program.
The Bash witness confirms by its exit code that it ran in place of the command.
The reference labels could follow the IEC protocol, so 2 annotators labeled both samples without their verdict, with $\kappa$ 0.79 and 0.89.

\paragraph{\textbf{External validity.}}
Changes on Linux happen before Bash parses the command, so the Linux corpora pass through the released command handling of 5 harnesses and 2 further shells.
We also search 203{,}710 public trajectories for the error of history expansion.
A repair tuned to the benchmark could fail in use, so \system ran in the production sessions and runs in the execution kernel of a commercial desktop agent.
Results on a few LLMs could reflect those LLMs rather than the path, so the scores cover 4 LLMs on \iecbench and 15 LLMs from 4 vendors on the fix and commit tasks.
\section{Related Work}
\label{sec:related}

\subsection{Evaluation and Failure Analysis of LLM Agents}
Agents complete tasks through tool calls when they develop software~\citep{yang2024sweagent,wang2025openhands,wang2024codeact,zhang2024autocoderover,bouzenia2025repairagent,bouzenia2025executionagent,yang2026aprsurvey,yang2025kgcompass}, operate applications~\citep{trivedi2024appworld,li2026toolathlon}, or use MCP servers~\citep{hou2026mcp,hasan2026mcp}.
Benchmarks of tool use score the selected call~\citep{patil2025bfcl,ye2025toolhop,basu2025nestful,yu2026wildtoolbench} or the state that an agent produces~\citep{trivedi2024appworld,barres2025tau2,jimenez2024swebench,zhang2025swebenchlive,luo2025mcpuniverse,wang2025mcpbench,li2026toolathlon,chen2026tuabench,pysklo2026agentdiff}, and ToolBench-X~\citep{tian2026toolbenchx} injects tool-environment hazards visible to the agent.
When a run fails, Who\&When~\citep{zhang2025whowhen} attributes the failure to an agent and a step, with an LLM judge that reads the failure log.
Causal Agent Replay~\citep{shah2026car} re-executes a recorded run from a step and intervenes on that step.
Further studies cover failure taxonomies~\citep{liu2025failures,majgaonkar2025behaviour,bouzenia2025trajectories,babu2026binding}, LLM-harness defects~\citep{chen2026arbugs,ning2026defects}, injected faults~\citep{tan2026agentchaos,seyedghorban2026observability,jia2026masfire}, record and replay~\citep{feng2025agentrr,chawla2026chronicle}, run instability~\citep{khachadourian2026dfah}, and the launches and engineering of widely-used harnesses~\citep{barbaste2026harness,liu2026divecc,jarmak2026reliable}.
Gupta et al.~\citep{gupta2026dissecting} identify an intent-execution gap and return structured execution feedback to the agent.
Qu et al.~\citep{qu2026integrity} argue for an integrity property from the user's intent to the execution.
Self-correction techniques such as Reflexion~\citep{shinn2023reflexion} depend on the feedback the agent receives~\citep{huang2024cannot,kamoi2024self,ding2024cycle}.

Trajectory-based analyses read the call and its result but not what a hop received, so a changed call is incorrectly attributed to the LLM.
The judges of Who\&When reach $\kappa$ of at most 0.04, compared with 0.77 for the IEC protocol.
With the same number of trials, \system passes significantly more items than Reflexion.

\subsection{Command Execution Paths}
Shell command generation~\citep{westenfelder2025llm,scriptsmith2025} evaluates the generated command.
Injection defenses~\citep{santosgrueiro2026cxi,wang2026agentspec}, CARE's pre-execution verification~\citep{liu2026care}, and security analyses of MCP~\citep{hou2026mcp,hasan2026mcp} ask whether a call may be executed.
Codex now tracks the quoting of PowerShell commands as an open issue.\footnote{\url{https://github.com/openai/codex/issues/11360}}
Su and Wassermann~\citep{su2006essence} detect injection by the changes untrusted input makes to a command's syntax, whereas the IEC protocol compares the structure that each hop received with the one the agent sent.
QuoteBench~\citep{li2026quotebench}, the closest study, records 1 hop, the wrapper, on 56 tasks against a single reference path.
This path bypasses 1 hop, goes through the others, and detects 37.2\% of the path-caused losses of \iecbench.

The injection defenses and security analyses guard one hop and ask whether a call may be executed, not whether the program received the call the agent sent.
QuoteBench, recording the wrapper alone, detects none of the 48 losses at the process interface that the IEC protocol names and \system recovers.
\section{Conclusion}
\label{sec:conclusion}

In this paper, we present the IEC protocol, which observes a call without executing it and names the first hop that changed it, and \system, which repairs the path at that hop.
In 47{,}828 shell calls within production sessions, Claude Code's Bash tool changes 12.0\% of the calls that carry code, escape sequences, or long text.
For 80.7\% of the calls whose backslashes are changed, the wrong action runs without any reported error.
All 10 measured harnesses change a call.
Trajectory-based judgment attributes 95.1\% of the production failures to the LLM, although the path caused more than half of them, whereas the IEC protocol agrees with human labels at $\kappa$ 0.77.
On \iecbench, the path raises the token cost per passed task 2.4 times (up to 12.3 times).
A hop that changes a call also hides the changes after it, and \system recovers 79.2\% of the failures with a changed call. These results show that a changed call comes from the launch, not the LLM.
A tool-call failure has to be analyzed and repaired across the hops of its path, and the harness establishes intent-execution correspondence independently of the LLM.

Future harnesses should therefore be designed and tested hop-by-hop, so that a correct call is executed as intended or refused, and benchmarks should report their launch and hold it fixed when comparing LLMs. The IEC protocol applies at every hop whose receiver has a parser, so the same measurement covers structured tools and MCP servers, and \iecbench grows with each change observed in use.
A repair that rewrites a call must leave the harness's permission rules judging the original call, as the mod does.
A benchmark that reports its launch and holds it fixed makes LLM scores comparable across harnesses.
We plan to extend the measurement to harnesses that type into a terminal and to launches that appear after this study.

\bibliographystyle{plainnat}
\bibliography{references}

\end{document}